\documentclass[sigconf]{acmart}

\AtBeginDocument{%
  \providecommand\BibTeX{{%
    \normalfont B\kern-0.5em{\scshape i\kern-0.25em b}\kern-0.8em\TeX}}}

\copyrightyear{2023} 
\acmYear{2023} 
\setcopyright{acmlicensed}\acmConference[CIKM '23]{Proceedings of the 32nd
ACM International Conference on Information and Knowledge
Management}{October 21--25, 2023}{Birmingham, United Kingdom}
\acmBooktitle{Proceedings of the 32nd ACM International Conference on
Information and Knowledge Management (CIKM '23), October 21--25, 2023,
Birmingham, United Kingdom}
\acmPrice{15.00}
\acmDOI{10.1145/3583780.3615045}
\acmISBN{979-8-4007-0124-5/23/10}

\usepackage{multirow}
\usepackage{subfigure}
\usepackage[ruled,vlined,linesnumbered]{algorithm2e}
\usepackage[capitalize,noabbrev]{cleveref}
\crefname{section}{Section}{Sections}
\crefname{theorem}{Theorem}{Theorems}
\crefname{lemma}{Lemma}{Lemmas}
\crefname{equation}{Eq.}{Equations}
\crefname{proposition}{Proposition}{Propositions}
\crefname{claim}{Claim}{Claims}
\crefname{appendix}{Appendix}{Appendices}
\crefname{algorithm}{Algorithm}{Algorithms}
\crefname{figure}{Figure}{Figures}
\crefname{table}{Table}{Tables}
\crefname{remark}{Remark}{Remarks}
\crefname{definition}{Def.}{Definitions}
\crefname{corollary}{Corollary}{Corollaries}

\usepackage{xspace}
\newcommand{\ours}[0]{\texttt{SAILOR}\xspace}

\begin{document}

\title{SAILOR: Structural Augmentation Based Tail Node Representation Learning}

\author{Jie Liao}
\orcid{0000-0002-1361-5325}
\affiliation{%
    \institution{Sun Yat-sen University}
    \city{Guangzhou}
    \country{China}}
\email{liaoj27@mail2.sysu.edu.cn}

\author{Jintang Li}
\orcid{0000-0002-6405-1531}
\affiliation{%
    \institution{Sun Yat-sen University}
    \city{Guangzhou}
    \country{China}}
\email{lijt55@mail2.sysu.edu.cn}

\author{Liang Chen}
\orcid{0009-0005-9682-0672}
\authornote{Liang Chen is the corresponding author.}
\affiliation{%
    \institution{Sun Yat-sen University}
    \city{Guangzhou}
    \country{China}}
\email{chenliang6@mail.sysu.edu.cn}

\author{Bingzhe Wu}
\affiliation{%
    \institution{Tecent AI Lab}
    \city{Shenzhen}
    \country{China}}
\email{wubingzhe@pku.edu.cn}

\author{Yatao Bian}
\affiliation{%
    \institution{Tecent AI Lab}
    \city{Shenzhen}
    \country{China}}
\email{yatao.bian@gmail.com}

\author{Zibin Zheng}
\affiliation{%
    \institution{Sun Yat-sen University}
    \city{Guangzhou}
    \country{China}}
\email{zhzibin@mail.sysu.edu.cn}

\renewcommand{\shortauthors}{Jie Liao et al.}

\begin{abstract}
    Graph neural networks (GNNs) have achieved state-of-the-art performance in representation learning for graphs recently.
    However, the effectiveness of GNNs, which capitalize on the key operation of message propagation, highly depends on the quality of the topology structure.
    Most of the graphs in real-world scenarios follow a long-tailed distribution on their node degrees, that is, a vast majority of the nodes in the graph are \emph{tail} nodes with only a few connected edges.
    GNNs produce inferior node representations for tail nodes due to the lack of sufficient structural information.
    In the pursuit of promoting the performance of GNNs for tail nodes, we explore how the deficiency of structural information deteriorates the performance of tail nodes and propose a general \underline{s}tructural \underline{a}ugmentation based ta\underline{il} n\underline{o}de \underline{r}epresentation learning framework, dubbed as \ours,
    which can jointly learn to augment the graph structure and extract more informative representations for tail nodes.
    Extensive experiments on six public benchmark datasets demonstrate that \ours outperforms the state-of-the-art methods for tail node representation learning.
\end{abstract}





\maketitle

\section{Introduction}
A majority of data on the Web can be represented in the form of graphs \cite{2010www-graphData1, 2004www-graphData2, CoraCiteseerPubmedData, AmazonCSData}.
For example, the association between the web pages can be modeled as a citation network \cite{2007arxiv-wikicsdata, 2021-wikipediaData} so that we can carry some network analysis \cite{2007-networkAnalysis1, 2007TKDD-networkAnalysis2} or graph algorithms \cite{2012VLDB-GraphLab, 2014OSDI-GraphX} on them to mine meaningful information for downstream applications.
, ranging from grouping pages by field to recommending web pages.
Among numerous graph data mining algorithms, graph neural networks (GNNs) \cite{2017ICLR-GCN, 2018ICLR-GAT, 2017NIPS-GraphSAGE, 2018ICML-JKnet, 2020KDD-AM-GCN} take the top spot in recent years.
GNNs are superior at learning representations for graphs and have been widely used in real-world scenarios, such as community detection on social media platforms \cite{2018aaai-social2,2019tkde-social3}, product recommendations on e-commerce systems \cite{2019ijcai-BundleRec1,kdd2022-BundleRec2}, and so forth.
In particular, GNNs rely on a global message propagation mechanism, which enables nodes to recursively gather information (\textit{i.e.}, node features like bags of words for web pages) from their neighbors.
Thus, the quality of the provided topological structure has a significant impact on the performance of GNNs.

\begin{figure}[t]
    \centering
    \subfigure[Long-tailed node distribution]{
        \includegraphics[width=0.22\textwidth]{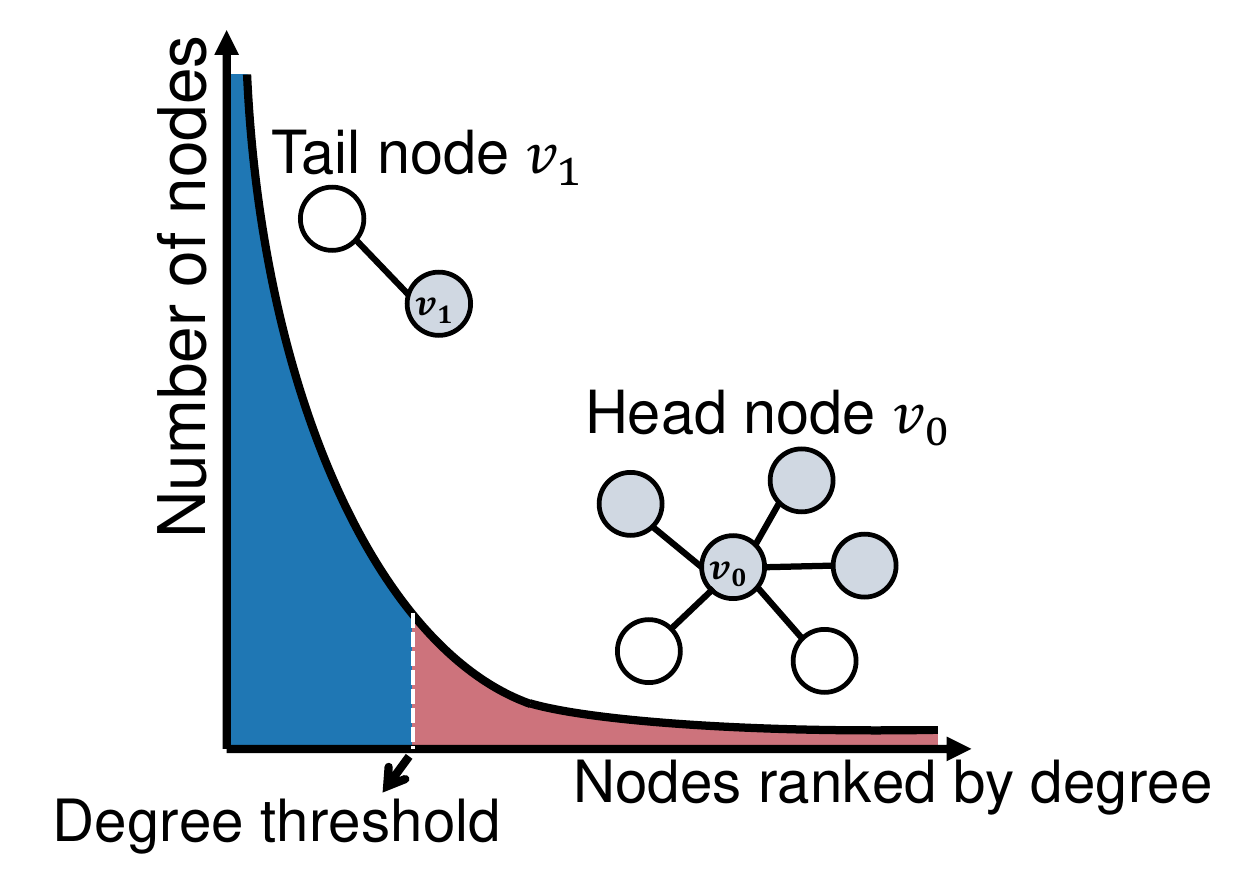}
        \label{fig:LongTailDistribution}
    }%
    \hfill
    \subfigure[Performance of GCN on nodes with different degrees]{
        \includegraphics[width=0.22\textwidth]{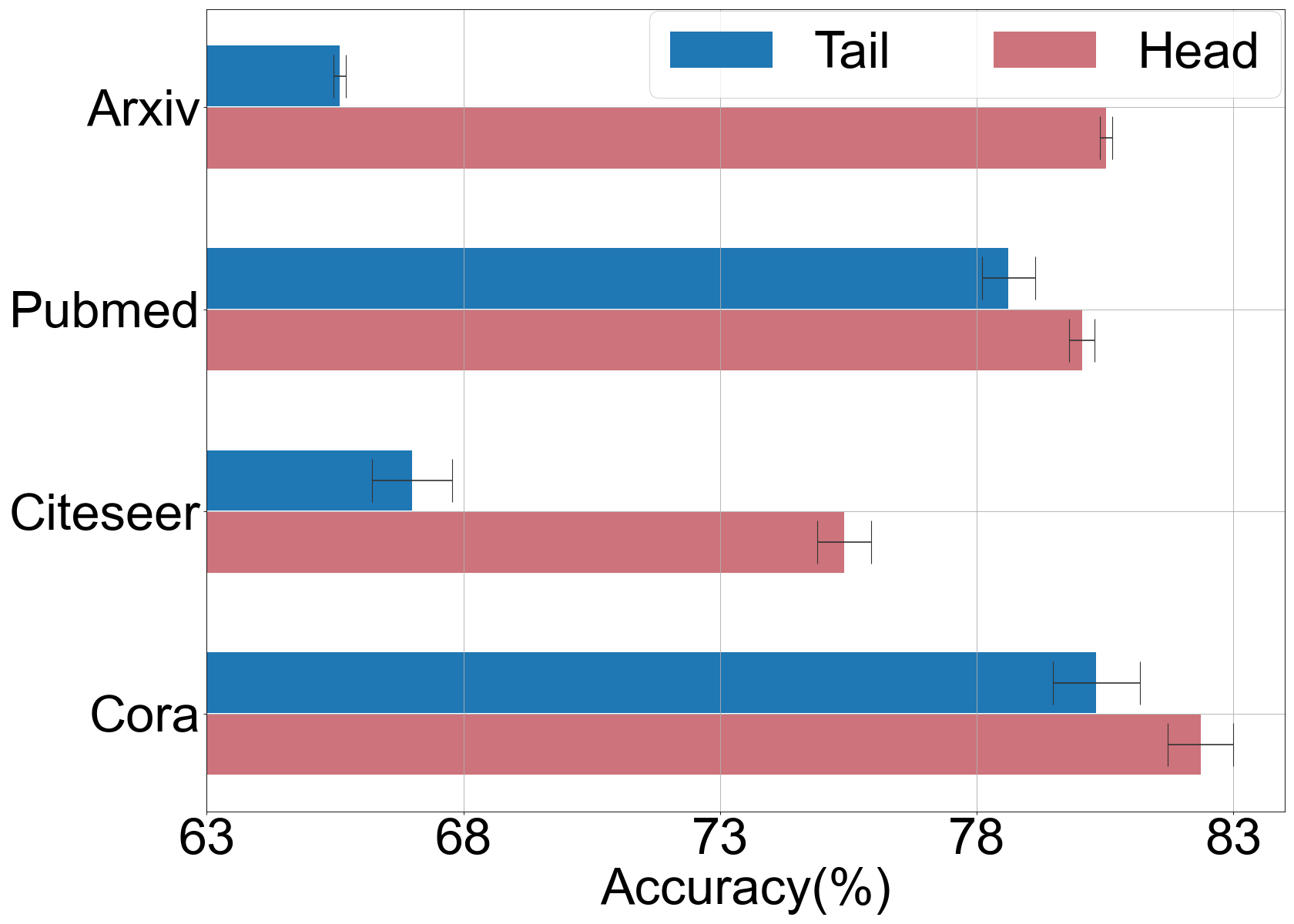}
        \label{fig:CiteseerPerformance}
    }
    \caption{Illustration of long-tailed node distribution. Empirical results show that GNNs perform inferior on tail nodes.}
    \label{fig:IllustrationTail}
\end{figure}

However, most graphs in real-world scenarios suffer from a severe imbalance of topology structure, \textit{i.e.}, they follow a long-tailed distribution on their node degrees.
Following previous work \cite{2020CIKM-meta-tail2vec, 2021KDD-TailGNN, 2022ICLR-ColdBrew}, we refer to the low-degree nodes as \textit{tail nodes} and the high-degree nodes as \textit{head nodes} in contrast.
\cref{fig:LongTailDistribution} illustrates that the bulk of the graph's nodes is tail nodes.
Particularly, the area of the blue block under the curve represents the total number of tail nodes, while the area of the red block beneath the curve indicates the total number of head nodes.
Since tail nodes are prevalent and constitute the majority of nodes in a graph, it is essential to focus on their performance in downstream tasks.
Unfortunately, most state-of-the-art GNNs do not attach special attention to tail nodes, and the deficiency in neighborhood information deteriorates their performance in learning representations for tail nodes, as demonstrated in \cref{fig:CiteseerPerformance}.
Therefore, in this work, we aim to improve the downstream task performance of tail nodes.

Recent research \cite{2019KDD-DEMO-Net, 2020CIKM-SL-DSGCN} has aimed to preserve degree-specific information for nodes to improve the performance of GNNs.
And the issue of degree-related bias in graph convolutional networks was first identified in \cite{2020CIKM-SL-DSGCN}.
The most recent work \cite{2022NIPS-GRADE} aims to alleviate structural unfairness and narrow the performance gap among nodes with different degrees.
However, none of these studies explicitly focus on improving representation learning for tail nodes.
More closely related studies \cite{2020CIKM-meta-tail2vec, 2021KDD-TailGNN} have formalized the long-tail problem on node degrees, that is, GNNs perform poorly on tail nodes (i.e., small-degree nodes), which make up the majority of nodes in a graph.
Both studies focus on improving tail node embeddings by identifying a shared neighborhood translation pattern among all nodes.
However, neither investigates the reason why a shortage of neighborhood knowledge causes tail nodes to perform worse.
These works \cite{2020CIKM-meta-tail2vec, 2021KDD-TailGNN} also introduce a setting where all head nodes are used for training and tail nodes for validation and testing.
This is practical because tail nodes often have limited ground-truth label information in real-world scenarios.
For example, an influencer’s occupation or gender is widely known and reliable, while that of a Twitter user with only a few followers may be uncertain.
The following state-of-the-art method \cite{2022ICLR-ColdBrew} leverages learning different feature transformation parameters, called \textit{structural embedding} in the work, for nodes with different degrees.
In this work, we follow \cite{2020CIKM-meta-tail2vec, 2021KDD-TailGNN, 2022ICLR-ColdBrew} and target the improvement of tail node representation learning.
The key difference between previous work and \ours is that we investigate how a shortage of structural information impairs the representation learning of tail nodes and propose to improve it from the perspective of feature diffusion.

\begin{figure}[t]
    \includegraphics[width=0.45\textwidth]{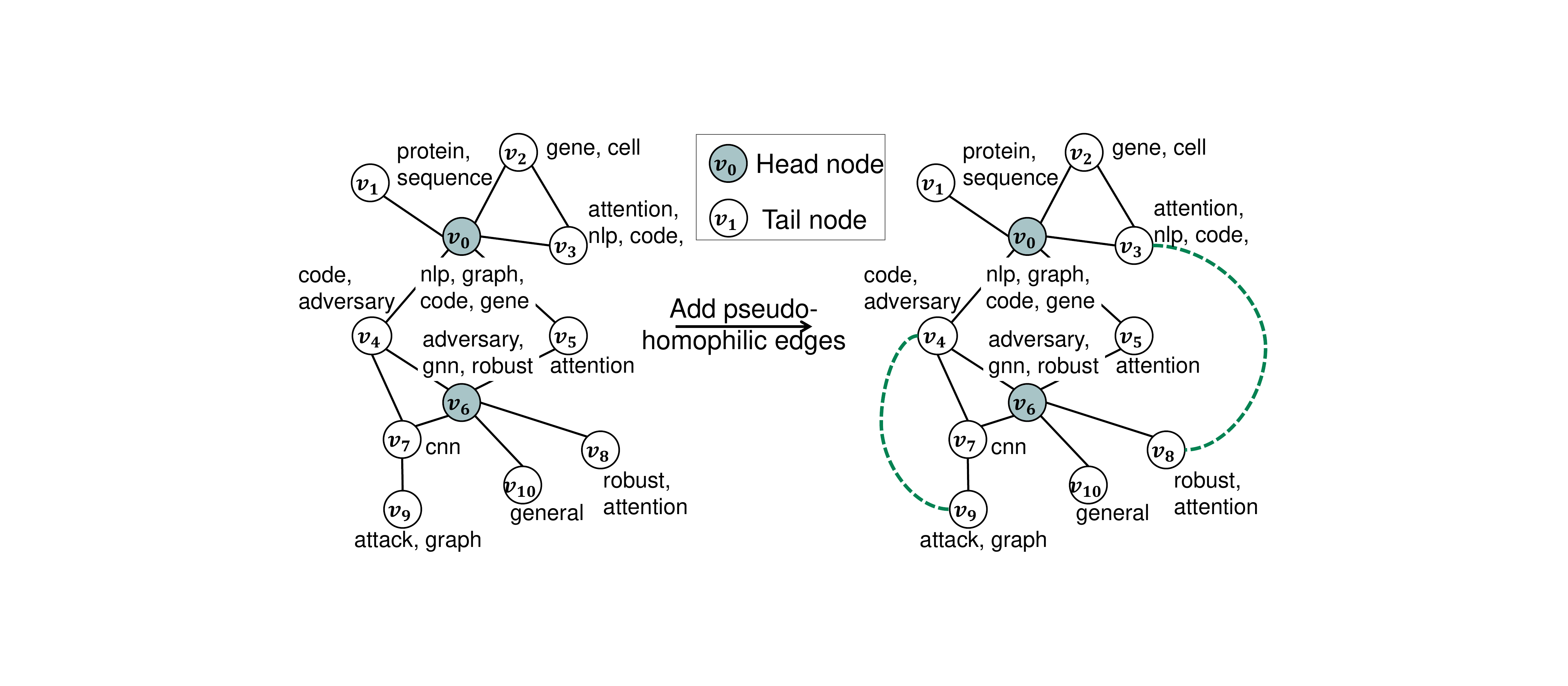}
    \caption{Illustration of adding pseudo-homophilic edges to tail nodes. More homophilic information can pass to tail nodes $v_8$ and $v_9$ via the added edges, which are presented as green dashed lines.}
    \Description{Illustration of pseudo-homophilic edges.}
    \label{fig:pseudoEdge}
\end{figure}

Specifically, we investigated the distribution of homophily among head and tail nodes.
In this work, \textit{homophily} describes the phenomenon that connected nodes in the graph tend to share the same label.
We discovered that due to the scarcity of neighbors,
tail nodes are more likely to exhibit \textit{total-heterophily},
meaning they do not share any labels with all their neighbors.
This finding is validated both theoretically and experimentally in \cref{sec:motivation}.
We propose that the lack of structural information leads to a deficiency of homophilic neighbors for tail nodes, which can hinder the effectiveness of the GNN’s message propagation mechanism.

Furthermore, it is known that GNNs propagate messages between neighboring nodes in each layer through two separate steps: feature diffusion and feature transformation.
While both head and tail nodes undergo the same feature transformation process with unified parameters, they aggregate different messages during the feature diffusion process.
Existing studies on tail node representation learning \cite{2020CIKM-meta-tail2vec, 2021KDD-TailGNN, 2022ICLR-ColdBrew} have focused on enhancing tail node embedding from the perspective of feature transformation.
However, they have overlooked the inherent deficiency of homophilic neighbors for tail nodes in terms of feature diffusion caused by the lack of structural information.

These observations motivate us to develop an effective tail structure augmentor that adds \textit{pseudo-homophilic edges}.
This reduces their total heterophily and improves the performance of GNNs for tail nodes from the feature diffusion aspect.
By \textit{homophilic edge}, we mean an edge that connects two nodes with the same label.
And we refer to the predicted homophilic edge by the augmentor as \textit{pseudo-homophilic edges}.
As illustrated in \cref{fig:pseudoEdge}, the pseudo-homophilic edges added to tail nodes $v_9$ and $v_8$ facilitate them to aggregate more homophily information from their direct surroundings and thus benefit the message-passing process.

The key challenge for the tail structure augmentor is \emph{how to locate the pseudo-homophilic edges for tail nodes without knowing the ground-truth labels}.
Furthermore, the developed model is expected to be versatile and applicable to various GNNs.
To address these challenges, we propose \ours, a general \underline{s}tructural \underline{a}ugmentation based ta\underline{il} n\underline{o}de \underline{r}epresentation learning framework, which jointly augments the structure for tail nodes and trains a GNN.
In summary, we make the following contributions:
\begin{itemize}
    \item We explore how the deficiency of structural information impairs the representation learning of tail nodes.
    \item We propose a general framework, \ours, that hinges on the key operation of adding pseudo-homophily edges to tail nodes to enable them to gain more task-relevant information from their immediate neighborhoods.
    \item We conduct extensive experiments on six public datasets and show that our proposed \ours outperforms state-of-the-art baselines on tail node representation learning.
\end{itemize}

\section{Related Work}
\subsection{Graph Representation Learning}
\subparagraph{\textbf{Graph neural networks.}}
Recent years have witnessed graph neural networks (GNNs) emerge as the most widely used methods for graph representation learning.
Assuming connected nodes tend to share similar attributes with each other, they \cite{2017ICLR-GCN, 2017NIPS-GraphSAGE, 2016NIPS-VGAE, 2018ICLR-GAT, 2018ICML-JKnet, 2019KDD-Cluster-GCN, 2020KDD-AM-GCN} usually follow a neighborhood aggregation scheme that enables each node to receive relevant neighborhood information and update its own representation.
Node classification is widely applied as the downstream task to assess the quality of the learned node representations.
Nevertheless, most state-of-the-art GNNs do not attach special attention to tail nodes and perform inferior on them.

\subparagraph{\textbf{Graph data augmentation.}}
Generally, graph data augmentation aims to obtain an optimal graph structure.
Existing techniques can be categorized into three folds: feature-wise augmentation, structure-wise augmentation, and label-wise augmentation \cite{2022arxiv-GraphDataAugmentation}.
Most relevant to our work is the literature devoted to augmenting the graph structure to obtain ideal connectivity.
Among these works, \cite{2020AAAI-AdaEdge, 2021AAAI-GAUG} employ specific strategies to eliminate noisy edges and introduce beneficial ones to attain an optimal graph structure.
\cite{2020KDD-ProGNN} devotes to obtaining a clean graph from a poisoned graph and defending against types of graph adversarial attacks \cite{2023TKDE-JintangAttack, arxiv-JintangAttackSurvey}.
However, our work focuses on a different problem, which is to enhance the representation learning for tail nodes.

\subsection{Tail Node Representation Learning}
Recent research \cite{2019KDD-DEMO-Net, 2020CIKM-SL-DSGCN} has aimed to preserve degree-specific information for nodes with varying degrees to improve the performance of GNNs.
Specifically, DEMO-Net \cite{2019KDD-DEMO-Net} employs a multi-task graph convolutional network to attain degree-specific embeddings for each node.
Following DEMO-Net \cite{2019KDD-DEMO-Net}, SL-DSGCN \cite{2020CIKM-SL-DSGCN} first explicitly points out the degree-bias issue and proposes the use of Bayesian neural networks to generate uncertainty scores for pseudo labels of unlabeled nodes, allowing for dynamic weighting of the training step size.
These models aim to improve the overall performance of GNNs and do not specifically enhance the learning of tail node representations.
The most recent work \cite{2022NIPS-GRADE} claims that graph contrastive learning can mitigate structural unfairness and develops a graph contrastive learning-based model called GRADE.
While GRADE \cite{2022NIPS-GRADE} aims to narrow the performance gap among nodes with different degrees and does not focus on boosting the representation learning for tail nodes, it fails to achieve competitive performance under the setting in \cite{2020CIKM-meta-tail2vec, 2021KDD-TailGNN}.

Meta-tail2vec \cite{2020CIKM-meta-tail2vec} proposes to treat the representation learning of each tail node as a meta-testing task and to construct the learning of head nodes as meta-training tasks.
Similarly, TailGNN \cite{2021KDD-TailGNN} aims to learn the translation pattern from a node to its neighborhood using tailored strategies.
This allows the proposed model to recover missing neighborhood information for tail nodes.
Cold Brew \cite{2022ICLR-ColdBrew} employs a teacher-student distillation approach, using a GNN as the teacher model and a multi-layer perceptron (MLP) as the student model.
It leverages node-wise structural embedding and a distillation scheme to complement missing neighborhood information for tail nodes.
In summary, all of these works provide innovative and effective strategies for enhancing tail node representation from the feature transformation aspect.
However, in terms of feature diffusion, they fail to consider the impact of the lack of structural information on the deficiency of homophilic neighbors for tail nodes.

\section{Preliminaries}
\label{preliminary}
\subparagraph{\textbf{Notation of graph}.}
Let $G=(A, X)$ be an undirected and unweighted graph, where $A \in \{0, 1\}^{N \times N}$ is the adjacency matrix of the graph and $X \in \mathbb{R}^{N \times F}$ denotes $F$-dimension feature vectors for all $N$ nodes.
Specifically, each entry $A_{ij}$ indicates whether the node $i$ is connected to node $j$, with a value of $1$ representing a connection and $0$ otherwise.
Additionally, the graph can also be represented as $G=(\mathcal{V}, \mathcal{E})$, where $\mathcal{V}$ denotes the set of nodes in the graph and $\mathcal{E}$ represents the set of edges.

\subparagraph{\textbf{Divison of head and tail nodes}.}
Following the Pareto principle \cite{ParetoPrinciple}, we first sort all nodes in ascending order according to their degrees.
We then select the top 80\% of nodes as tail nodes, denoted as $\mathcal{V}_{tail}$, and the remaining nodes as head nodes, denoted as $\mathcal{V}_{head}$.
In practice, we calculate the total number of tail nodes while retrieving the degrees in ascending order.
If the current number of tail nodes exceeds 80\% of all nodes, we stop retrieving and set the corresponding degree as the degree threshold.

\subparagraph{\textbf{Problem Formulation}.}
Given a graph $G=(A, X)$, the goal of node representation learning is to identify a function $g: A, X \rightarrow Z$ that maps each node to a low-dimensional vector.
$Z\in\mathbb{R}^{N\times d}$ denotes the output embedding matrix of all nodes, where $d$ is the dimension of the embedding vectors.
In a node classification task, $d$ is often set to $C$, which corresponds to the number of node classes in the given dataset.
In this paper, our focus is on improving the quality of tail node representations, that is, $\{Z_v\vert v \in \mathcal{V}_{tail}\}$.

\section{The heterophily of tail nodes}
\label{sec:motivation}

\begin{figure}[t]
    \centering
    \subfigure[Comparison of performance on total-heterophilic nodes versus other nodes.]{
        \includegraphics[width=0.22\textwidth]{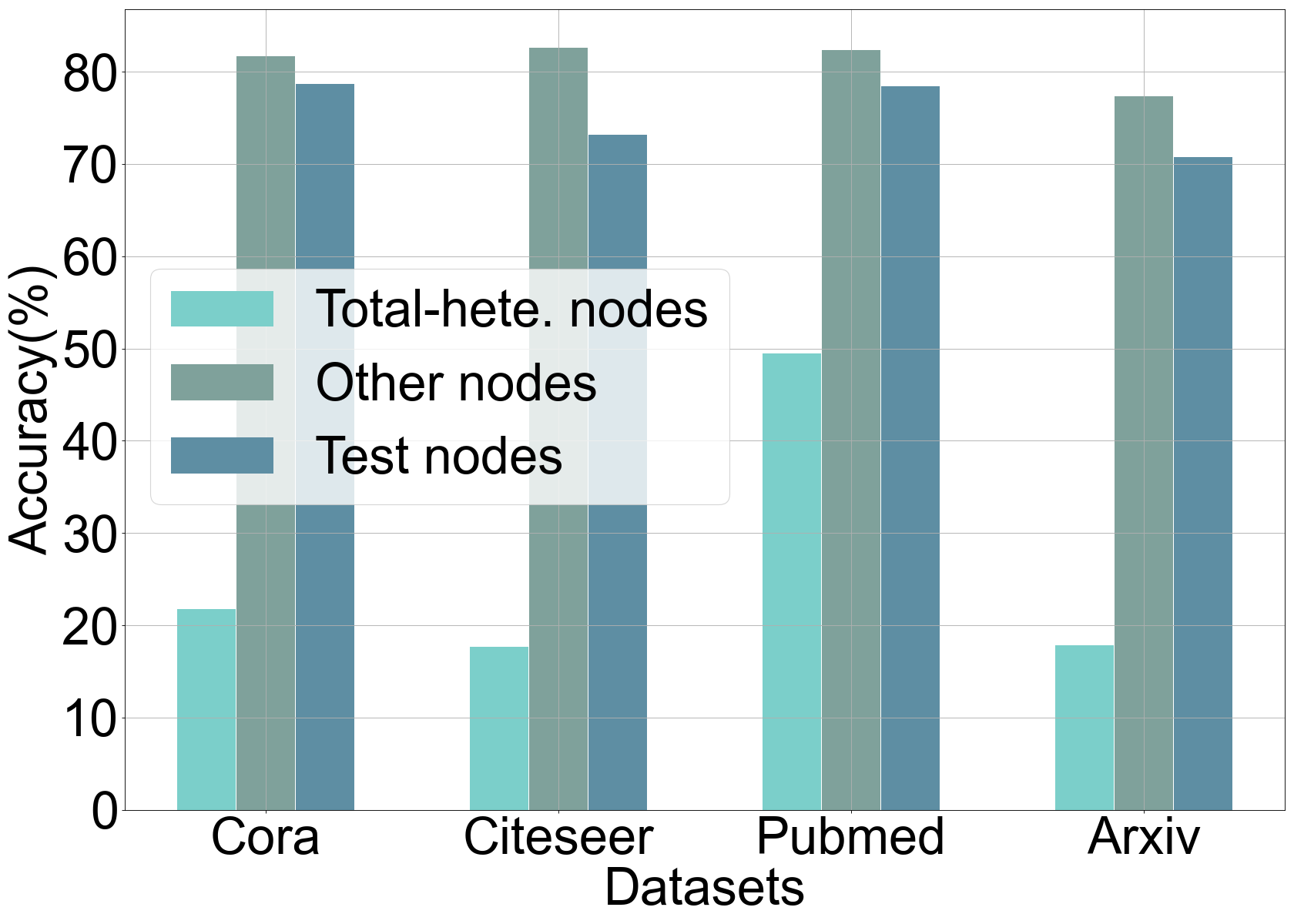}
        \label{fig:heterophily-acc}
    }
    \hfill
    \subfigure[Homophily distribution.]{
        \includegraphics[width=0.22\textwidth]{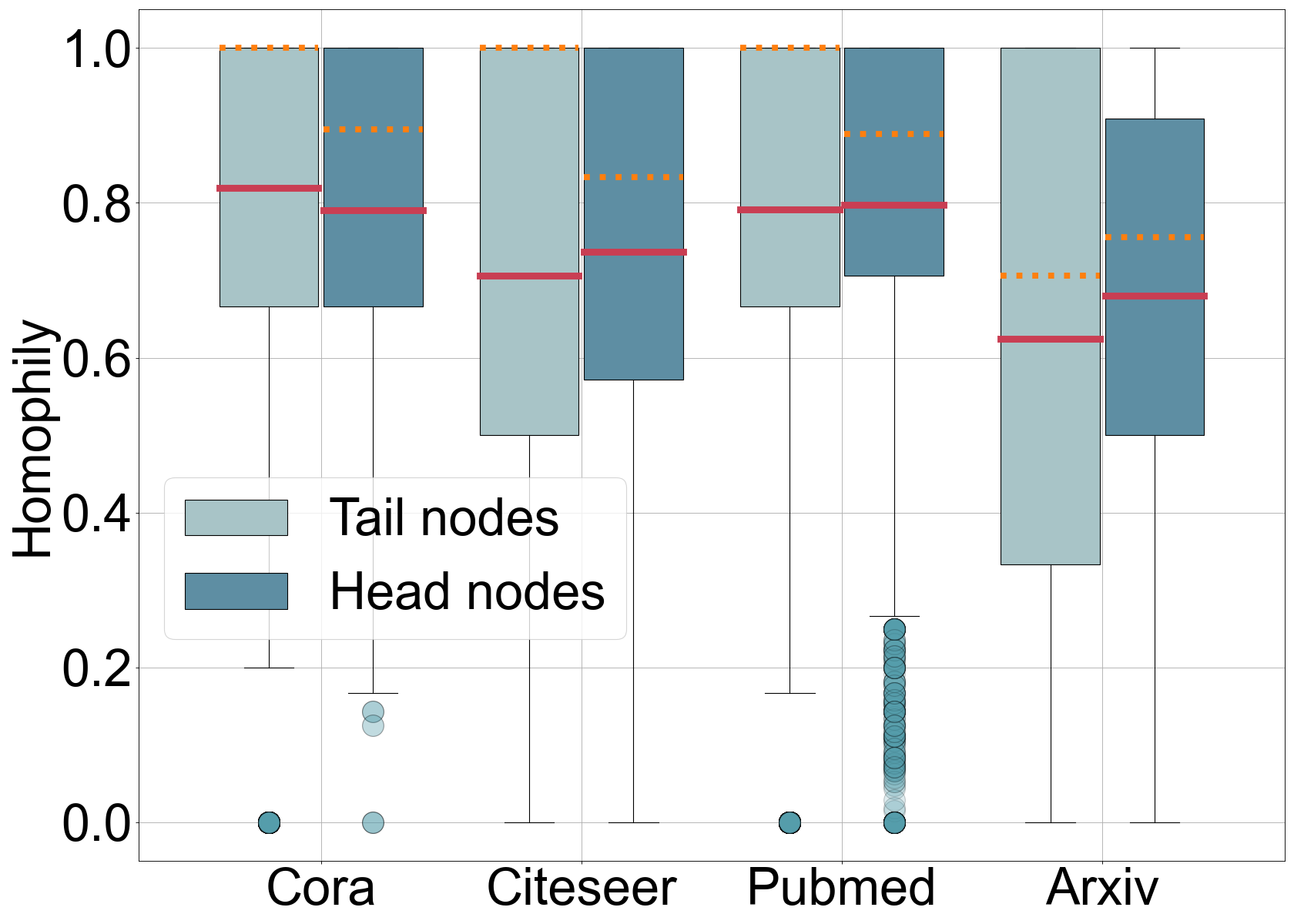}
        \label{fig:homophily-box}
    }
    \caption{Illustration of heterophily of tail nodes.}
\end{figure}

We notice that while nodes aggregate different messages during the feature diffusion process, they all share the same feature transformation process with unified model parameters.
For instance, head nodes have more neighbors than tail nodes, allowing them to gather more information from their immediate surroundings.
This could be a contributing factor to why GNNs perform less effectively on tail nodes than on head nodes.

Recent works \cite{2020NIPS-homophily1, 2021ICLR-homophily2, 2020ICLR-homophily3} have shown that the homophily of the underlying graph, which permits nodes to obtain more task-relevant information from neighbors, has a significant impact on the performance of GNNs \cite{2017ICLR-GCN, 2018ICLR-GAT, 2017NIPS-GraphSAGE, 2018ICML-JKnet}.
In other words, if a node shares no labels with any of its neighbors, GNNs will fail to learn high-quality representation for it because it completely violates the homophily assumption.
We call this kind of nodes total-heterophilic nodes in this work.
We present the average node classification accuracy of a 2-layer GCN \cite{2017ICLR-GCN} on total-heterophilic nodes in \cref{fig:heterophily-acc}, and the results are in line with our expectation.
It demonstrates that the performance of GNNs on total-heterophilic nodes is far inferior to other nodes in the test set.

Assuming that node labels are randomly and uniformly distributed, the probability of a head node becoming a total-heterophil-ic node (denoted as $\mathcal{P}_{h}$)  is significantly lower than that of a tail node (denoted as $\mathcal{P}_{t}$).
Formally, given that $D_{h}$ and $D_{t}$ represent the degree of a head and tail node respectively, $\mathcal{P}_{h}$ can be expressed as $(\frac{C-1}{C})^{D_{h}}$, while $\mathcal{P}_{t}$ is $(\frac{C-1}{C})^{D_{t}}$.
Since $D_{h}$ is much greater than $D_{t}$, it follows that $\mathcal{P}_{h}$ is significantly lower than $\mathcal{P}_{t}$.
This suggests that in a homophily graph, head nodes can gather more homophilic information from their direct neighbors to enhance their learned representations.
In contrast, tail nodes are more likely to suffer from a lack of homophilic neighbors or even total-heterophily due to their limited number of neighbors, resulting in poorer performance than head nodes.
To verify this hypothesis, we examine the homophily of head and tail nodes in detail.

We calculate the homophily \cite{2020ICLR-GeomGCN} for each head node and tail node individually and present the results in \cref{fig:homophily-box}.
Specifically, the red solid line represents the mean homophily of tail nodes and head nodes across four public datasets.
The orange dotted line indicates the median homophily value, while the light-blue dots denote outliers.
The color intensity of the dots in \cref{fig:homophily-box} corresponds to the number of outliers, with a higher number of outliers resulting in a darker dot color.
As shown in \cref{fig:homophily-box}, the average homophily of head nodes is higher than that of tail nodes in three datasets: Citeseer, Pubmed, and Arxiv.
Although the node homophily of tail nodes is higher in Cora, they still suffer from more severe total-heterophily than head nodes, which has a greater impact on their inferior downstream-task performance.
In addition, \cref{fig:homophily-box} also indicates that tail nodes have more heterophilic neighbors in two ways:
in the Citeseer, Pubmed, and Arxiv datasets, the box for tail nodes is longer, meaning that the lower quartile homophily is lower;
in the Cora and Pubmed datasets, tail nodes have a higher number of total-heterophilic neighbors.

To better illustrate the difference in the proportion of total-heterophilic nodes between head and tail nodes, we provide \cref{tab:heterophily}.
From \cref{tab:heterophily}, we can make two observations.
Firstly, the majority of total-heterophilic nodes in the graph are tail nodes.
For instance, out of the 147 total-heterophilic nodes in Cora, 143 are tail nodes.
Secondly, the proportion of total-heterophilic nodes in tail nodes is significantly greater than that in head nodes.
In Cora, for instance, 6.89\% of tail nodes are total-heterophilic, compared to only 0.97\% of head nodes.
These findings suggest that tail nodes aggregate lower-quality information during the neighborhood aggregation process compared to head nodes, due to the higher proportion of total-heterophilic nodes.
This introduces noise into the training of model parameters, resulting in an inferior performance of GNNs for these nodes.
This motivates us to develop a model that can enhance the structural information for tail nodes and improve the performance of GNNs for these nodes.

\begin{table}[t]
    \centering
    \caption{The amount and proportion of total-heterophilic nodes in head nodes, tail nodes, respectively.}
    \begin{tabular}{c|c|c|c|c|c}
        \toprule
        \multicolumn{2}{c|}{}          & \textbf{Cora} & \textbf{Citeseer} & \textbf{Pubmed} & \textbf{Arxiv}           \\
        \midrule
        \multirow{2}{*}{\textbf{Head}} & amount        & 4                 & 16              & 50             & 331     \\
                                       & proportion    & 0.97\%            & 4.78\%          & 1.38\%         & 1.00\%  \\
        \midrule
        \multirow{2}{*}{\textbf{Tail}} & amount        & 143               & 308             & 2,471          & 20,468  \\
                                       & proportion    & 6.89\%            & 17.25\%         & 15.37\%        & 15.01\% \\
        \bottomrule
    \end{tabular}
    \label{tab:heterophily}
\end{table}

\section{The Proposed Framework: SAILOR}

\begin{figure}[t]
    \includegraphics[width=0.85\linewidth]{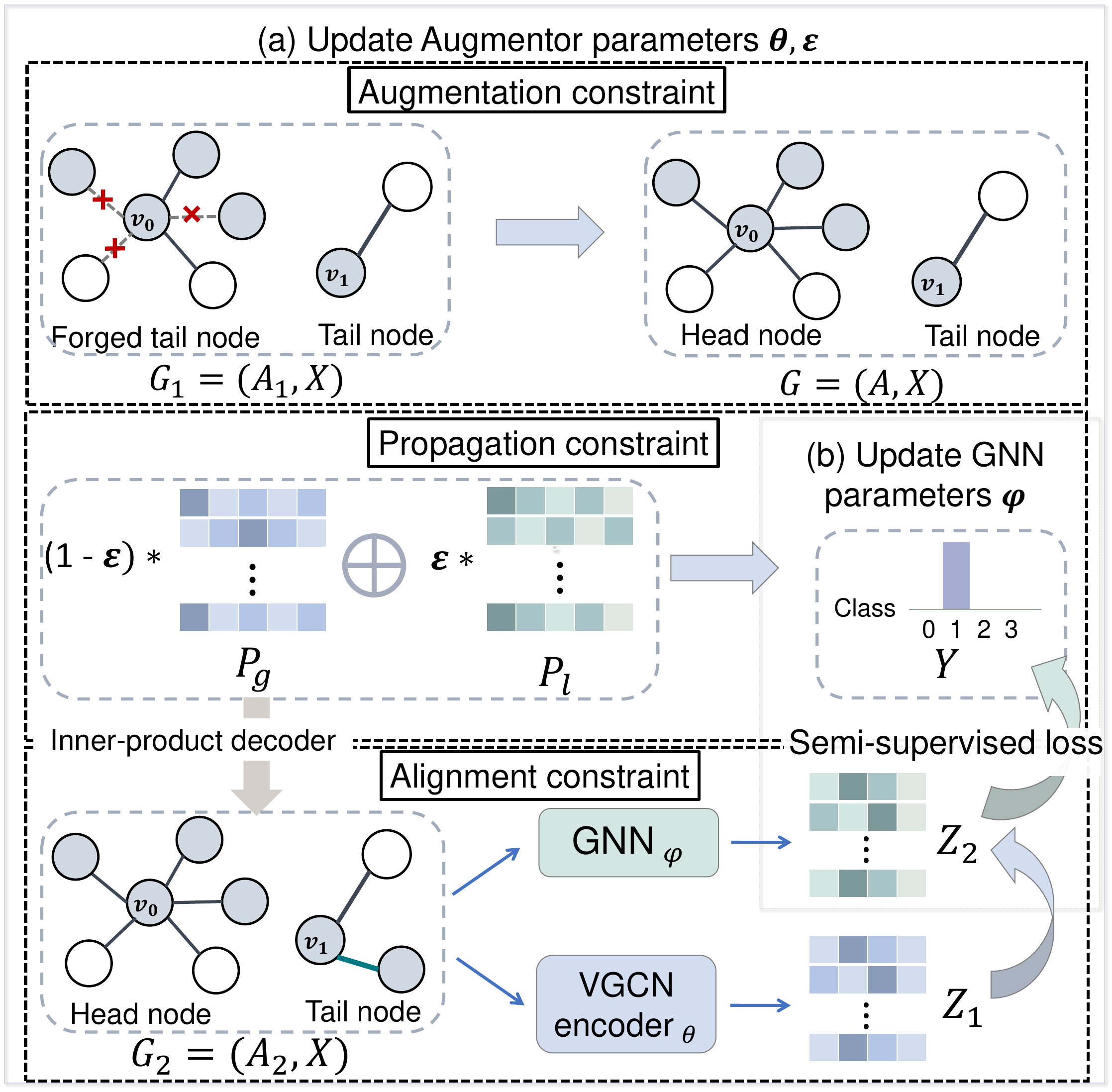}
    \caption{Overview of \ours, which exploits a tail structure augmentor to add pseudo-homophilic edges to each tail node. This allows tail nodes to aggregate more homophilic information from their immediate neighborhoodds. By utilizing the augmented graph $G_2$ as input for training, the GNN can learn parameters that more accurately reflect the actual feature transformation pattern.
    }
    \Description{framework.}
    \label{fig:framework}
\end{figure}

In this section, we present a detailed description of our proposed framework, \ours.
The key idea behind \ours is the use of a tail structure augmentor to add pseudo-homophilic edges to each tail node.
This allows tail nodes to aggregate more task-relevant information from their neighborhoods.
By utilizing the augmented graph as input for training, the GNN can learn parameters that more accurately fit the actual feature transformation pattern.
As illustrated in \cref{fig:framework}, \ours consists of two main components: the Tail Structure Augmentor (Augmentor) and the Graph Neural Network (GNN).
(a) The Augmentor is trained with the augmentation, propagation, and alignment loss to update its parameters and generates an augmented graph $G_2$.
(b) The GNN takes $G_2$ as input and is trained using a semi-supervised loss to update its feature transformation parameters $\phi$.
It learns node embeddings $Z_2$ to fine-tune the Augmentor’s parameters $\theta$ under the constraint of the alignment loss.
As a result, these two components can enhance each other and be trained jointly.

\subsection{Graph Neural Networks}
\label{subsec:GNNs}
GNNs rely on a message propagation mechanism to recursively update node representations.
This mechanism drives the performance of GNNs to be highly dependent on the homophily of the underlying graph.
In particular, GNNs propagate messages between neighboring nodes in each layer, which can be regarded as two separate steps.
Firstly, every node aggregates features from its neighborhood and we call this step \textit{feature diffusion}.
Mathematically,
\begin{equation}
    \label{eq:feature_diffusion}
    \hat{H}^{(l)} = A H^{(l)},
\end{equation}
where $H^{(l)}$ denotes the hidden representations of the graph at the $l$-th layer and the initial $H^{(0)}$ is the node feature matrix $X$.
In fact, the aggregation scheme can vary across different GNNs \cite{2017NIPS-GraphSAGE, 2017ICLR-GCN, 2018ICLR-GAT, 2018ICML-JKnet}, such as mean pooling and max pooling.
Besides, the adjacent matrix can also be normalized as $\widetilde{A}=\hat{D}^{-\frac{1}{2}}\hat{A}\hat{D}^{-\frac{1}{2}}$, where $\hat{A}=A+I$ and $\hat{D}$ is the degree matrix of $\hat{A}$.
For illustration purposes, we use \cref{eq:feature_diffusion}, which is just one example that essentially employs sum pooling.

Subsequently, all nodes transform their aggregated representations with unifying model parameters $\phi$.
This step can be referred to as \textit{feature transformation}.
Formally,
\begin{equation}
    \label{eq:feature_transform}
    H^{(l+1)} = g_\phi^{(l)}(\hat{H}^{(l)}) = \sigma(\hat{H}^{(l)}W_{\phi}^{(l)}),
\end{equation}
where $W_{\phi}^{(l)}$ denotes the trainable parameters at layer $l$ and $\sigma$ is the nonlinear activation function.
We take \cref{eq:feature_transform} for the convenience of illustration.
In fact, the transformation form can vary.
For example, it can also be written as $g_\phi^{(l)}(\hat{H}^{(l)})$$ = \hat{H}^{(l)}W_{\phi}^{(l)}+b_{\phi}^{(l)}$, where $b_{\phi}^{(l)}$ are learnable parameters.

            It is now evident that both head and tail nodes follow the same feature transformation pattern with unified model parameters, as shown in \cref{eq:feature_transform}.
            However, they aggregate different information during the feature diffusion process, as shown in \cref{eq:feature_diffusion}.
            As discussed in \cref{sec:motivation}, the performance of tail nodes deteriorates due to a lack of sufficient homophilic neighbors for training.

            Thus, our proposed framework \ours trains the GNN with the augmented graph $G_2=(A_2, X)$ which is generated by a tail structure augmentor.
            Since there are fewer total-heterophilic tail nodes in the augmented graph, it can facilitate the training of the GNN.
            The above process can be formulated as follows:
            \begin{equation}
                \begin{aligned}
                                  & \hat{Z_2}^{(l)} = A_2 Z_2^{(l)}, \\
                    Z_2^{(l+1)} = & g_{\phi}^{(l)}(\hat{Z_2}^{(l)})
                    = \sigma(\hat{Z_2}^{(l)}W_{\phi}^{(l)}),
                \end{aligned}
            \end{equation}
            where $\sigma$ is the activation function, $\phi=\{ W_{\phi}^{(l)} \, \vert \, l\in\{0,1,2,\dots,\,$$L_2-1\} \}$ are trainable parameters, and $L_2$ is the total amount of layers of the GNN.
        When $l=0$, the hidden representation is initialized by $X$, i.e., $Z_2^{(0)}=X$. Besides, we refer to the output of the last layer as $Z_2$.

        Following common practice, we train the GNN with the semi-supervised loss, which is expressed as follows:
        \begin{equation}
            \label{eq:L_t}
            \begin{aligned}
                \mathcal{L}_{sup} = \text{CE}(Z_2, Y),
            \end{aligned}
        \end{equation}
        where $\text{CE}(\cdot)$ is the cross-entropy function and $Y$ represents the ground-truth labels of nodes in the training set.

        \subsection{Tail Structure Augmentor}
        \label{subsec:tail structure augmentor}
        As suggested by previous work \cite{2014ICLR-VAE, 2016NIPS-VGAE}, learning the distribution of hidden embeddings instead of learning individual hidden embeddings can help train more robust parameters.
        Therefore, we deploy a Variational Graph Convolution Network (VGCN) based encoder with learnable parameters $\theta$ as the backbone architecture for the tail structure augmentor.
        Moreover, we design three constraint strategies to optimize the augmentor.

        \begin{figure}[t]
            \includegraphics[width=0.85\linewidth]{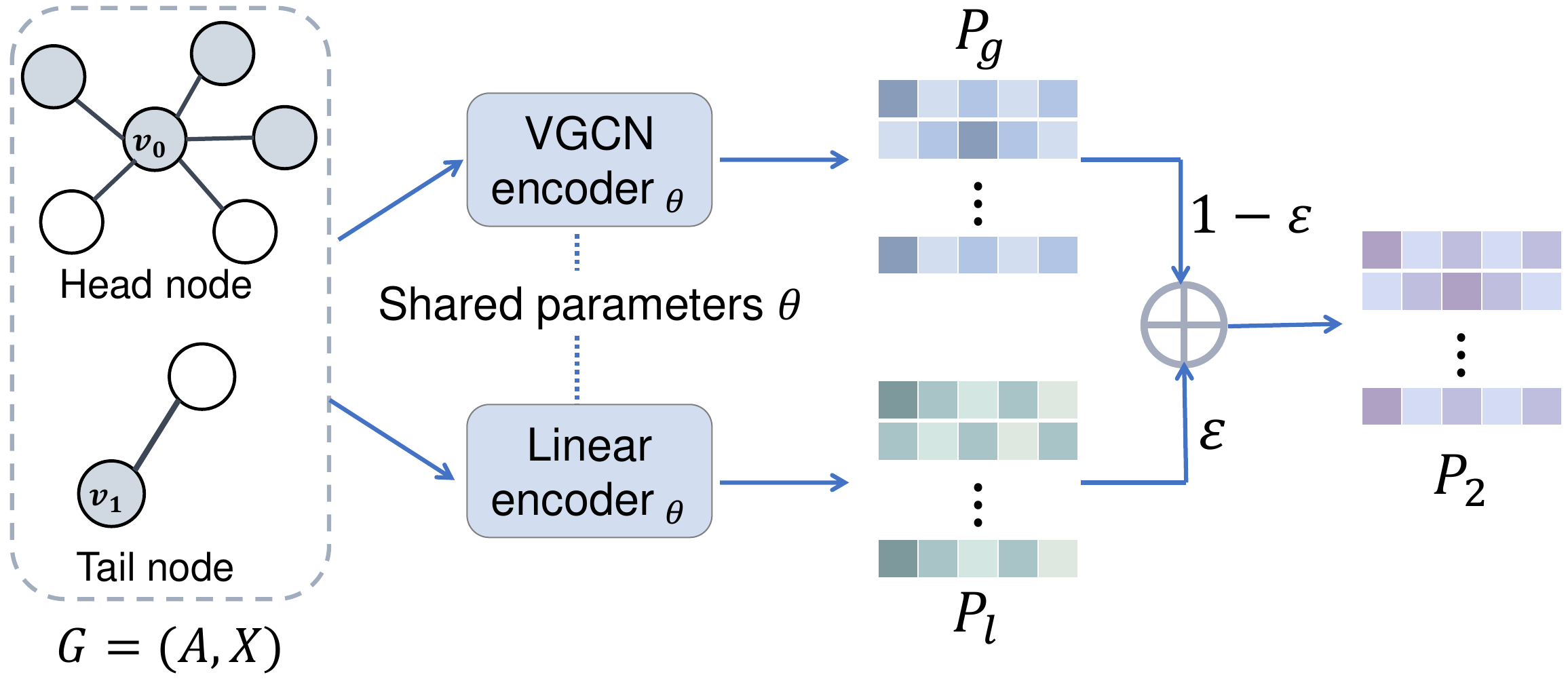}
            \caption{The VGCN encoder and linear encoder share the same feature transformation parameters $\theta$. In contrast to the VGCN encoder, the linear encoder does not perform message propagation on nodes but instead only performs feature transformation. $\varepsilon$ are trainable parameters that adaptively adjust the importance of the graph structure.
            }
            \Description{propgation_loss.}
            \label{fig:augmentor}
        \end{figure}

        \subparagraph{\textbf{Augmentation constraint}.}
        Since the Variational Graph Auto-Encoder (VGAE) \cite{2016NIPS-VGAE} is commonly used for link prediction, one might presume that it could be directly adopted to add edges for tail nodes.
        However, this intuitive solution is not feasible and does not perform well due to two challenges.
        Firstly, directly using VGAE for neighbor generation can make the model lazy as its optimal situation is to recover the original structure. In this case, the inferior performance of tail nodes remains.
        Secondly, in view that there are heterophilic edges in the original graph structure, this direct approach may introduce noise to the node classification task.

        Towards the challenges, we propose to go beyond VGAE by training a mapping function from forged tail nodes to head nodes, allowing us to learn how to map tail nodes to structurally augmented tail nodes.
        Specifically, we randomly drop edges for each head node and turn them into forged tail nodes, inspired by Tail-GNN \cite{2021KDD-TailGNN}.
        In this work, we utilize the forged tail nodes for training to replicate the circumstance of insufficient neighborhood information, which can also help mitigate the degree distribution shift problem from the training set to the test set under the dataset partitioning setting in \cite{2020CIKM-meta-tail2vec, 2021KDD-TailGNN}.
        In contrast, Tail-GNN follows the idea of GAN (Generative Adversarial Nets) \cite{2014NIPS-GAN} and exploits the forged tail nodes to train a smarter discriminator.

        Formally, given the original graph $G=(A, X)$, we randomly drop a certain proportion, called $\Delta$, of edges for each head node and get a new adjacency matrix $A_1$.
        It is evident that the original graph $G$ is denser than $G_1=(A_1, X)$,
        in which even the head nodes may lack sufficient neighborhood information.
        With $G_1=(A_1, X)$ as input, the VGCN encoder $f_{\theta}(\cdot)$ conducts neighborhood aggregation for each node at each layer.
        It can be formulated as:
        \begin{equation}
            \label{eq:A_1X}
            P_1^{(l+1)} = f_{\theta}^{(l)}(A_1, P_1^{(l)}) = \sigma(A_1 P_1^{(l)}W_{\theta}^{(l)}),
        \end{equation}
        where
    $W_{\theta}^{(l)}$ are trainable parameters of the VGCN encoder at layer $l$,
    $P_1^{(l)}$ represents the hidden representations at layer $l$,
    $P_1^{(0)}$ is the input feature matrix $X$,
        and $\sigma$ denotes the activation function.

        Furthermore, we use $f_{\theta_\mu}(\cdot)$ and $f_{\theta_\sigma}(\cdot)$ to transform the output representations of the last hidden layer into the mean representation $\boldsymbol{\mu}$ and standard deviation representation $\boldsymbol{\sigma}$, respectively.
        By mapping node features to the entire latent space instead of a single point, the parameters $\phi$ become more robust and help to produce more representative node embeddings.
        We then sample from the learned distribution to obtain the embeddings $P_1$.
        To enable backward propagation of gradients during the sampling process, we employ the reparameterization trick as described in \cite{2016NIPS-VGAE}.
        Formally,
        \begin{equation}
            \label{eq:Z_1}
            \begin{aligned}
                \boldsymbol{\mu} =    & f_{\theta_\mu} (P_1^{(L_1)}) = P_1^{(L_1)} W_{\theta}^{\mu},              \\
                \boldsymbol{\sigma} = & f_{\theta_\sigma} (P_1^{(L_1)}) = P_1^{(L_1)} W_{\theta}^{\sigma},        \\
                                      & P_1 = \boldsymbol{\mu} + \boldsymbol{\sigma} \odot \boldsymbol{\epsilon},
            \end{aligned}
        \end{equation}
        where $\boldsymbol{\epsilon} \in \mathcal{N}(0, I)$ is the noise variable sampled from Gaussian distribution,
    $L_1$ represents the number of hidden layers of the VGCN encoder,
        and $\boldsymbol{\mu}, \boldsymbol{\sigma} \in \mathbb{R}^{N\times d}$.

        Subsequently, we adopt the inner product and the activation function $\text{sigmoid}(\cdot)$ to decode $P_1$ and obtain $A^{\prime}_1$.
        Formally,
        \begin{equation}
            \label{eq:decode_P_1}
            \begin{aligned}
                A^{\prime}_1 = \text{sigmoid}(P_1 P_1^T).
            \end{aligned}
        \end{equation}
        We expect the tail structure augmentor to restore missing edges for the forged tail nodes.
        As a result, the original graph structure $A$ serves as the supervised signal for $A_1^{\prime}$.
        Moreover, we intend to reduce the Kullback Leibler (KL) divergence \cite{KL-divergence} between the distribution of embeddings $P_1$ and a Gaussian distribution, following \cite{2016NIPS-VGAE}.
        In summary, the augmentation constraint objective for the tail structure augmentor can be formulated as:
        \begin{equation}
            \label{eq:L_r}
            \begin{aligned}
                \mathcal{L}_{aug} = \text{BCE}(A_1^{\prime}, A) + \text{KL}\left[q(P_1 \vert A_1, X) \Vert p(P_1)\right],
            \end{aligned}
        \end{equation}
        where $p(P_1) = \prod \limits_{i=1}^N \mathcal{N}(\mathbf{p}_1^i\vert0, I)$ is a Gaussian prior of node representations,
        and $q(P_1 \vert A_1, X) = \prod \limits_{i=1}^N \mathcal{N}(\mathbf{p}_1^i\vert \boldsymbol{\mu}, \text{diag}(\boldsymbol{\sigma}^2))$ is the learned distribution.
    $\text{BCE}(\cdot)$ is the binary cross entropy function.

        \subparagraph{\textbf{Propagation constraint}.}
        As observed in \cref{sec:motivation}, even in a homophily graph, the graph structure of some nodes may be unreliable and introduce noise to the training of the GNN.
        To mitigate the negative impact of noisy graph structures of some nodes, we adopt a learnable parameter $\varepsilon \in \mathbb{R}^N$ to automatically adjust the importance of the graph structure for node representation learning.
        Specifically, we utilize both a VGCN encoder and a linear encoder on the input graph to derive embeddings $P_g$ and $P_l$, respectively.
        It is important to note that the linear encoder shares parameters $\theta$ with the VGCN encoder.
        However, in contrast to the VGCN encoder, the linear encoder does not perform message passing on nodes but instead relies solely on parameters $\theta$ for feature transformation.
        Let $h_{\theta}(\cdot)$ denote the linear encoder, the above procedure can be formulated as:
        \begin{equation}
            \label{eq:get_Pl_Pg}
            \begin{aligned}
                P_g^{(l+1)} = f_{\theta}^{(l)}(A, X) = \sigma(AP_g^{(l)}W_{\theta}^{(l)}),                \\
                P_l^{(l+1)} = h_{\theta}^{(l)}(A, X) = h_{\theta}(X) = \sigma(P_l^{(l)}W_{\theta}^{(l)}), \\
            \end{aligned}
        \end{equation}
        where $P_g^{(l)}$ and $P_l^{(l)}$ represent the hidden representations of the VGCN encoder and the linear encoder at layer $l$, respectively.
        When $l=0$, both $P_g^{(0)}$ and $P_l^{(0)}$ are initialized as the input feature matrix $X$. We refer to the output of the last layer as $P_g$ and $P_l$ respectively.

        As illustrated in \cref{fig:augmentor}, $\varepsilon$ serves as the importance weight of $P_l$, while $1-\varepsilon$ serves as the one of $P_g$.
        Formally, we obtain $P_2$ as follows:
        \begin{equation}
            \label{eq:get_P2}
            \begin{aligned}
                P_2 = \varepsilon*P_l + (1-\varepsilon)P_g.
            \end{aligned}
        \end{equation}
        By adaptively adjusting the weight of the embeddings learned by the two encoders through $\varepsilon$, we can dynamically fine-tune the importance of the graph structure for learning node representation $P_2$.
        The ground-truth labels of the training set serve as a supervised signal for $P_2$, allowing us to optimize parameters $\varepsilon$ and $\theta$.
        This optimization process enables the augmentor to learn better parameters $\theta$, thereby facilitating the generation of higher-quality augmented graphs.
        Mathematically, we formulate the propagation constraint objective for the tail structure augmentor as follows:
        \begin{equation}
            \label{eq:propagation_loss}
            \begin{aligned}
                \mathcal{L}_{p} = \text{CE}(P_2, Y),
            \end{aligned}
        \end{equation}
        where $\text{CE}(\cdot)$ is the cross-entropy function and $Y$ represents the ground-truth labels of nodes in the training set.

        \subparagraph{\textbf{Alignment constraint}.}
        To generate the augmented graph $G_2 = (A_2, X)$, the tail structure augmentor employs the inner product to decode $P_2$.
        To identify the most likely existing edges among nodes, we apply the activation function $\text{softmax}(\cdot)$ on each row of the decoded matrix to derive a peak distribution instead of using the activation function $\text{sigmoid}(\cdot)$.
        We then adopt a Bernoulli-sampling procedure to obtain binary graph structure data, where $1$ indicates an existing edge and $0$ otherwise.
        Note that the sampled edges are the desired pseudo-homophily edges to be added to tail nodes.
        This process can be expressed mathematically as follows:
        \begin{equation}
            \label{eq:get_A2}
            \begin{aligned}
                A^{\prime}_2 & = \text{softmax}(P_2 P_2^T),                       \\
                A_2          & = \text{Clamp}(\text{BernSamp}(A^{\prime}_2) + A),
            \end{aligned}
        \end{equation}
        where $\text{BernSamp}(\cdot)$ denotes the Bernoulli-sampling procedure and $\text{Clamp}(\cdot)$ represents a function that clamps values to $[0,1]$.

        The GNN mentioned in \cref{subsec:GNNs} is then trained on $G_2$ to obtain refined model parameters $\phi$.
        Based on $\phi$, the parameters $\theta$ of the augmentor can be further fine-tuned, which in return allows the augmentor to produce higher-quality augmented graphs.
        Specifically, given the augmented graph $G_2$ as input to the VGCN encoder $f_{\theta}(\cdot)$, it derives node representations $Z_1$ as follows:
        \begin{equation}
            \label{eq:get_Z1}
            \begin{aligned}
                Z_1^{(l+1)} = f_{\theta}^{(l)}(A_2, Z_1^{(l)}) & = \sigma(A_2 Z_1^{(l)}W_{\theta}^{(l)}), \quad l=0,\dots,L_1-1        \\
                \boldsymbol{\mu}                               & = f_{\theta_\mu} (Z_1^{(L_1)}) = Z_1^{(L_1)} W_{\theta}^{\mu},        \\
                \boldsymbol{\sigma}                            & = f_{\theta_\sigma} (Z_1^{(L_1)}) = Z_1^{(L_1)} W_{\theta}^{\sigma},  \\
                Z_1                                            & = \boldsymbol{\mu} + \boldsymbol{\sigma} \odot \boldsymbol{\epsilon}.
            \end{aligned}
        \end{equation}

        We then exploit an alignment objective function to narrow the gap between $Z_1$ and $Z_2$, thereby fine-tuning the parameters $\theta$ under the guidance of $\phi$.
        Specifically, we employ KL divergence to measure the information loss from the learned distribution $r(Z_1 \vert \theta)$ to $t(Z_2 \vert \phi)$ and take $t(Z_2 \vert \phi)$ as the prior.
        Formally,
        \begin{equation}
            \label{eq:alignment_loss}
            \begin{aligned}
                \mathcal{L}_{ali} = KL\left[r(Z_1 \vert \theta) \Vert t(Z_2 \vert \phi)\right].
            \end{aligned}
        \end{equation}

        \begin{algorithm}[t]
            \caption{The training process of \ours.}
            \label{alg:train}
            \KwIn{Graph $G=(A, X) = (\mathcal{V}, \mathcal{E})$,
                tail nodes $\mathcal{V}_{tail}$, batch size $B$, learning rate $lr_g$, $lr_a$}
            \KwOut{GNN parameters $\phi$}
            Initialize model parameters $\theta, \varepsilon, \phi$\;
            \While{not converged}{
                $i=0$ and the edge set of the augmented graph $\mathcal{E}_2=\mathcal{E}$\;
                \tcp{Add pseudo-homophilic edges to tail nodes.}
                \While{$ i < \vert \mathcal{V}_{tail} \vert$}{
                    batch = $\mathcal{V}_{tail}$[$i$:$i+B$]\;
                    $A^{\prime}_{2b} = \text{softmax}(P_2[batch]\,P_2^T)$\;
                    $\mathcal{E'} = \text{Nonzero}(\text{BernSamp}(A^{\prime}_{2b}))$\;
                    $\mathcal{E}_2 = \mathcal{E}_2 \bigcup \mathcal{E'}$\;
                    $i = i + B$\;
                }
                \tcp{Update the GNN.}
                $L_{sup}$ $\leftarrow$ $\text{CE}(Z_2, Y)$\;
                $\phi$ \, $\leftarrow$ \, $\phi - lr_g\,\frac{\partial \, \alpha L_{sup}}{\partial \, \phi}$ \;

                \tcp{Update the augmentor.}

                $L_{\theta,\varepsilon}$ $\leftarrow$ $\beta \mathcal{L}_{aug}+\eta\mathcal{L}_{p}+\delta\mathcal{L}_{ali}$\;
                $\theta$ \, $\leftarrow$ \, $\theta - lr_a\,\frac{\partial \, L_{\theta,\varepsilon}}{\partial \, \theta}$ \;
                $\varepsilon$ \, $\leftarrow$ \, $\varepsilon - lr_a\,\frac{\partial \, L_{\theta,\varepsilon}}{\partial \, \varepsilon}$ \;
            }
            \Return{$\phi$}
        \end{algorithm}

        \subsection{Optimization of \ours}
        \label{subsec:optimization}

        In summary, the overall objective function of \ours can be formulated as follows:
        \begin{equation}
            \begin{aligned}
                \mathop{\arg\min}\limits_{\theta, \varepsilon, \phi} & \mathcal{L}
                = \mathcal{L}_{\theta, \varepsilon}
                + \alpha\mathcal{L}_{\phi},                                                                   \\
                \text{where} \quad
                                                                     & \mathcal{L}_{\phi}= \mathcal{L}_{sup}, \\
                \text{and} \quad
                                                                     & \mathcal{L}_{\theta, \varepsilon}
                = \beta \mathcal{L}_{aug}
                + \eta \mathcal{L}_{p}
                + \delta \mathcal{L}_{ali},
            \end{aligned}
        \end{equation}
        where the predefined parameter $\alpha$ controls the impact of the GNN on the overall optimization process.
    $\beta$, $\eta$, and $\delta$ govern the relative weights of the three constraints for the tail structure augmentor.
        We train the tail structure augmentor and the GNN concurrently to optimize $\mathcal{L}$.
        The augmented graph $G_2$ is generated by the tail structure augmentor and used to train the GNN.
        In return, the GNN derives node representations $Z_2$, which serve as the prior to fine-tune the tail structure augmentor.
        As a result, these two components enhance each other and are trained jointly.

        To provide a clear explanation of our method, we detail the training algorithm as in \cref{alg:train}.
        Although decoding the node representations to the graph structure can lead to a space complexity of $O(N^2)$,
        it is easy to achieve an acceptable space complexity of $O(B^2)$ by batch-processing. Here $B$ denotes the number of nodes in a batch.
        Additionally, we deploy the commonly used early-stopping mechanism to accelerate the model training and prevent overfitting.

        \begin{table}[t]
            \centering
            \caption{Summary of datasets. “DegTh.” denotes the degree threshold that split the nodes into $80\%$ tail nodes and $20\%$ head nodes, as stated in \cref{preliminary}. "\#Feat." denotes the dimension of each node feature vector.}
            \scalebox{.92}{
                \begin{tabular}{cccccc}
                    \toprule
                     & \#Node    & \#Edge & \#Feat. & \#Class
                     & DegTh.                                 \\
                    \midrule
                    \textbf{Chameleon} \cite{2021-wikipediaData}
                     & 2,277                                  
                     & 62,742                                 
                     & 2,325                                  
                     & 5                                      
                     & 33                                     \\ 
                    \textbf{Squirrel} \cite{2021-wikipediaData}
                     & 5,201                                  
                     & 396,706                                
                     & 2,089                                  
                     & 5                                      
                     & 123                                    \\       
                    \textbf{Cora} \cite{CoraCiteseerPubmedData}
                     & 2,485                                  
                     & 10,138                                 
                     & 1,433                                  
                     & 7                                      
                     & 5                                      \\
                    \textbf{Citeseer} \cite{CoraCiteseerPubmedData}
                     & 2,120                                  
                     & 7,358                                  
                     & 3,703                                  
                     & 6                                      
                     & 5                                      \\
                    \textbf{Pubmed} \cite{CoraCiteseerPubmedData}
                     & 19,717                                 
                     & 88,648                                 
                     & 500                                    
                     & 3                                      
                     & 6                                      \\
                    \textbf{Arxiv} \cite{2020NIPS-OGBData}
                     & 169,343                                
                     & 2,315,598                              
                     & 128                                    
                     & 40                                     
                     & 17                                     \\
                    \bottomrule
                \end{tabular}
            }
            \label{tab:Data}
        \end{table}

        \section{Experiments}

        \begin{table*}[t]
            \centering
            \caption{Evaluation of tail node classification using GCN as the base model. The best results are shown in bold and the runners-up are underlined. "OOM" denotes that the model runs out of memory.}
            \begin{tabular}{ccccccccc}
                \toprule
                 & \textbf{Metric (\%)}
                 & \textbf{GCN}
                 & \textbf{DEMO-Net}
                 & \textbf{Meta-tail2vec}
                 & \textbf{Tail-GCN}
                 & \textbf{Cold Brew}
                 & \textbf{GRADE}
                 & \textbf{SAILOR}            \\
                \midrule

                \multirow{2}{*}{\textbf{Chameleon}}
                 & Accuracy
                 & 36.66$\pm$2.27             
                 & 32.23$\pm$0.86             
                 & 19.08$\pm$2.57             
                 & 31.06$\pm$1.21             
                 & \underline{40.01}$\pm$0.39 
                 & 33.81$\pm$0.94             
                 & \textbf{40.10}$\pm$0.90    \\
                 & Weighted-F1
                 & 34.24$\pm$2.81             
                 & 29.09$\pm$1.11             
                 & 13.00$\pm$1.89             
                 & 25.77$\pm$3.49             
                 & \underline{37.20}$\pm$0.73 
                 & 29.19$\pm$1.94             
                 & \textbf{38.56}$\pm$1.00    \\
                \midrule

                \multirow{2}{*}{\textbf{Squirrel}}
                 & Accuracy
                 & 30.04$\pm$1.55             
                 & 20.82$\pm$0.56             
                 & 19.04$\pm$0.19             
                 & 20.38$\pm$0.61             
                 & \underline{31.75}$\pm$1.02 
                 & 23.05$\pm$0.71             
                 & \textbf{32.79}$\pm$0.46    \\
                 & Weighted-F1
                 & \underline{26.79}$\pm$2.29 
                 & 16.27$\pm$0.71             
                 & 9.97$\pm$0.78              
                 & 13.91$\pm$0.92             
                 & 26.56$\pm$1.89             
                 & 18.84$\pm$0.99             
                 & \textbf{27.48}$\pm$1.02    \\
                \midrule

                \multirow{2}{*}{\textbf{Cora}}
                 & Accuracy
                 & 85.16$\pm$0.49             
                 & 84.43$\pm$0.72             
                 & 77.58$\pm$0.48             
                 & 84.27$\pm$0.79             
                 & \underline{86.13}$\pm$0.12 
                 & 82.77$\pm$0.85             
                 & \textbf{86.92}$\pm$0.50    \\
                 & Weighted-F1
                 & 85.20$\pm$0.50             
                 & 84.57$\pm$0.69             
                 & 77.63$\pm$0.50             
                 & 84.16$\pm$0.89             
                 & \underline{86.11}$\pm$0.13 
                 & 82.75$\pm$0.86             
                 & \textbf{86.93}$\pm$0.47    \\
                \midrule

                \multirow{2}{*}{\textbf{Citeseer}}
                 & Accuracy
                 & \underline{72.06}$\pm$0.48 
                 & 71.39$\pm$0.46             
                 & 65.59$\pm$0.99             
                 & 71.81$\pm$1.18             
                 & 70.76$\pm$0.39             
                 & 66.65$\pm$2.25             
                 & \textbf{74.30}$\pm$0.47    \\
                 & Weighted-F1
                 & \underline{71.08}$\pm$0.53 
                 & 70.29$\pm$0.67             
                 & 63.99$\pm$1.06             
                 & 70.18$\pm$1.03             
                 & 69.43$\pm$0.42             
                 & 64.54$\pm$1.95             
                 & \textbf{72.22}$\pm$0.50    \\
                \midrule

                \multirow{2}{*}{\textbf{Pubmed}}
                 & Accuracy
                 & \underline{85.85}$\pm$0.15 
                 & 83.69$\pm$0.15             
                 & 74.20$\pm$1.42             
                 & 85.09$\pm$0.29             
                 & 83.96$\pm$0.13             
                 & 81.57$\pm$0.94             
                 & \textbf{86.34}$\pm$0.17    \\
                 & Weighted-F1
                 & \underline{85.78}$\pm$0.16 
                 & 83.61$\pm$0.15             
                 & 73.66$\pm$1.59             
                 & 85.06$\pm$0.29             
                 & 83.92$\pm$0.15             
                 & 81.56$\pm$0.92             
                 & \textbf{86.31}$\pm$0.17    \\
                \midrule

                \multirow{2}{*}{\textbf{Arxiv}}
                 & Accuracy
                 & \underline{62.26}$\pm$0.23 
                 & >3days                     
                 & 23.60$\pm$7.63             
                 & 48.13$\pm$0.48             
                 & 61.19$\pm$0.33             
                 & OOM                        
                 & \textbf{63.52}$\pm$0.12    \\
                 & Weighted-F1
                 & \underline{60.54}$\pm$0.25 
                 & >3days                     
                 & 14.67$\pm$6.82             
                 & 41.24$\pm$0.46             
                 & 59.08$\pm$0.40             
                 & OOM                        
                 & \textbf{62.09}$\pm$0.17    \\

                \bottomrule
            \end{tabular}
            \label{tab:Result_gcn}
        \end{table*}

        \begin{table*}[t]
            \centering
            \caption{Evaluation of tail node classification using other GNNs as the base model. "Acc." denotes accuracy(\%) and "WF1." denotes weighted-f1 score (\%). "OOM" denotes that the model runs out of memory.}
            \begin{tabular}{c|cccc|cccc}
                \toprule
                 & \multicolumn{2}{c}{\textbf{GAT}}
                 & \multicolumn{2}{c|}{\textbf{SAILOR}}
                 & \multicolumn{2}{c}{\textbf{GraphSAGE}}
                 & \multicolumn{2}{c}{\textbf{SAILOR}}                         \\
                 & \textbf{Acc. (\%)}                     & \textbf{WF1. (\%)}
                 & \textbf{Acc. (\%)}                     & \textbf{WF1. (\%)}
                 & \textbf{Acc. (\%)}                     & \textbf{WF1. (\%)}
                 & \textbf{Acc. (\%)}                     & \textbf{WF1. (\%)} \\
                \hline
                {\textbf{Chameleon}}
                 & 31.12$\pm$1.76                                              
                 & 25.90$\pm$2.54                                              
                 & \textbf{36.25}$\pm$1.62                                     
                 & \textbf{32.77}$\pm$3.06                                     
                 & 32.04$\pm$3.68                                              
                 & 29.12$\pm$6.35                                              
                 & \textbf{37.38}$\pm$1.45                                     
                 & \textbf{35.29}$\pm$1.62                                     \\

                    {\textbf{Squirrel}}
                 & 20.81$\pm$3.38                                              
                 & 10.29$\pm$4.69                                              
                 & \textbf{26.00}$\pm$1.54                                     
                 & \textbf{20.09}$\pm$2.45                                     
                 & 26.47$\pm$2.12                                              
                 & 19.30$\pm$3.70                                              
                 & \textbf{30.72}$\pm$1.43                                     
                 & \textbf{23.97}$\pm$2.91                                     \\

                    {\textbf{Cora}}
                 & 85.31$\pm$0.55                                              
                 & 85.34$\pm$0.60                                              
                 & \textbf{86.50}$\pm$0.42                                     
                 & \textbf{86.52}$\pm$0.45                                     
                 & 85.39$\pm$0.46                                              
                 & 85.41$\pm$0.48                                              
                 & \textbf{86.48}$\pm$0.40                                     
                 & \textbf{86.52}$\pm$0.39                                     \\

                    {\textbf{Citeseer}}
                 & 71.55$\pm$0.77                                              
                 & 70.08$\pm$0.88                                              
                 & \textbf{73.72}$\pm$0.99                                     
                 & \textbf{72.30}$\pm$1.01                                     
                 & 71.20$\pm$0.56                                              
                 & 69.71$\pm$0.68                                              
                 & \textbf{74.39}$\pm$0.61                                     
                 & \textbf{72.69}$\pm$0.63                                     \\

                    {\textbf{Pubmed}}
                 & 85.28$\pm$0.15                                              
                 & 85.25$\pm$0.15                                              
                 & \textbf{85.39}$\pm$0.19                                     
                 & \textbf{85.36}$\pm$0.19                                     
                 & 84.62$\pm$0.14                                              
                 & 84.58$\pm$0.14                                              
                 & \textbf{84.96}$\pm$0.19                                     
                 & \textbf{84.92}$\pm$0.19                                     \\

                    {\textbf{Arxiv}}
                 & OOM                                                         
                 & OOM                                                         
                 & OOM                                                         
                 & OOM                                                         
                 & 62.44$\pm$0.23                                              
                 & 60.61$\pm$0.32                                              
                 & \textbf{63.85}$\pm$0.13                                     
                 & \textbf{62.08}$\pm$0.17                                     \\

                \bottomrule
            \end{tabular}
            \label{tab:Result_otherGNNs}
        \end{table*}

        In this section, we evaluate the effectiveness of \ours through comprehensive experiments.
        Firstly, we compare \ours to state-of-the-art tail node representation learning methods using two evaluation metrics on six public datasets.
        The results demonstrate that \ours outperforms the baselines.
        Secondly, we compare the homophily distribution of all nodes in the original graph to that in the augmented graph, which shows that the pseudo-homophilic edges are located as expected.
        Thirdly, we compare \ours with its four variants to examine the importance of different components and explore its sensitivity to different hyper-parameters.
        Lastly, we align the dataset partitioning setting with that of GCN \cite{2017ICLR-GCN} to further confirm the effectiveness of \ours and show that the performance of head nodes is not compromised by \ours.

        \subsection{Experimental Setups}
        \label{subsec:experiment setups}

        \subparagraph{\textbf{Datasets and Settings.}}
        We evaluate \ours with a total of six benchmark datasets, as summarized in \cref{tab:Data}.
        For each dataset, we extract the largest connected component and processed edges as undirected.
        We follow previous work \cite{2020CIKM-meta-tail2vec, 2021KDD-TailGNN} by using all head nodes for training and splitting the tail nodes into validation and test sets with a ratio of 1:4.
        All baselines are initialized with the settings provided in their respective open-source codes.
        We further fine-tune these settings to improve their performance.
        Experiments are conducted ten times using different random seeds, and we present the average results along with standard deviations.
        The source code is available at https://github.com/Jie-Re/SAILOR.git.

        \subparagraph{\textbf{Baselines.}}
        We compare \ours with classic GNNs and state-of-the-art tail node representation learning methods, which include:
        \begin{itemize}
            \item \textbf{GCN} \cite{2017ICLR-GCN}:
                  It is a widely used graph neural network.

            \item \textbf{DEMO-Net} \cite{2019KDD-DEMO-Net}:
                  It is a degree-aware graph neural network.
                  We employ GCN as its base architecture.

            \item \textbf{Meta-tail2vec} \cite{2020CIKM-meta-tail2vec}:
                  It applies the meta-learning framework MAML \cite{2017ICML-MAML} and learns to refine tail node embeddings.

            \item \textbf{Tail-GCN} \cite{2021KDD-TailGNN}:
                  It exploits an end-to-end framework to recover missing neighborhood features for tail nodes.
                  We employ GCN as its base architecture for a fair comparison.

            \item \textbf{Cold Brew} \cite{2022ICLR-ColdBrew}:
                  It devotes to addressing the strict cold start problem.
                  Since the MLP student of Cold Brew is inferior to the teacher GNN on tail nodes, we report the performance of the latter and employ GCN as its base architecture.

            \item \textbf{GRADE} \cite{2022NIPS-GRADE}: It is a graph contrastive learning-based meth-od aiming to narrow the performance gap among nodes with different degrees.
        \end{itemize}

        \subsection{Long-tailed Node Classification}
        We conduct the node classification task on six public datasets to assess the effectiveness of \ours.
        We use accuracy as our evaluation metric.
        Additionally, previous works \cite{2020CIKM-meta-tail2vec, 2021KDD-TailGNN} have also adopted the micro-F score as an evaluation metric.
        However, the micro-F score used by Tail-GNN \cite{2021KDD-TailGNN} is calculated by ignoring the largest class, which differs from the standard definition.
        In fact, the commonly used Micro-F1 score is mathematically equivalent to accuracy when there is one ground-truth label for each data point \cite{2021AAAI-GAUG}.
        To maintain the consideration of class imbalance in the evaluation metric of previous work \cite{2020CIKM-meta-tail2vec, 2021KDD-TailGNN}, while adhering to the standard definition, we adopt the Weighted-F1 score from sklearn.

        We employ the well-established GCN as the base model for all the GNN-based approaches to ensure a fair comparison.
        To examine the generalizability of \ours for other GNN architectures, we also show how it performs on GAT and GraphSAGE.

        \subparagraph{\textbf{Using GCN as the base model.}}
        \cref{tab:Result_gcn} displays the tail node classification results.
        The superior performance of \ours demonstrates that it learns a more representative feature transformation pattern and extracts high-quality tail node representations.
        From these results, we can draw two conclusions.
        Firstly, \ours achieves state-of-the-art performance compared to existing tail node representation learning baselines,
        suggesting that jointly augmenting the graph structure and refining the node representations can enhance the model.
        Secondly, \ours outperforms GCN, its underlying GNN architecture, indicating that the pseudo-homophilic edges
        generated by the tail structure augmentor
        are useful and facilitate the training of the GNN.

        \subparagraph{\textbf{Using other GNNs as base models.}}
        \ours is a versatile framework that can be applied to various GNNs.
        To validate this claim, we further implemented GAT and GraphSAGE within the framework and present their results in \cref{tab:Result_otherGNNs}.
        These results also demonstrate the superiority of \ours over the vanilla GNN models.

        \subsection{Importance of Structure Augmentation}
\begin{figure}[t]
            \centering
            \subfigure[Cora]{
                \label{subfig:augment_homophily_cora}
                \includegraphics[width=0.22\textwidth]{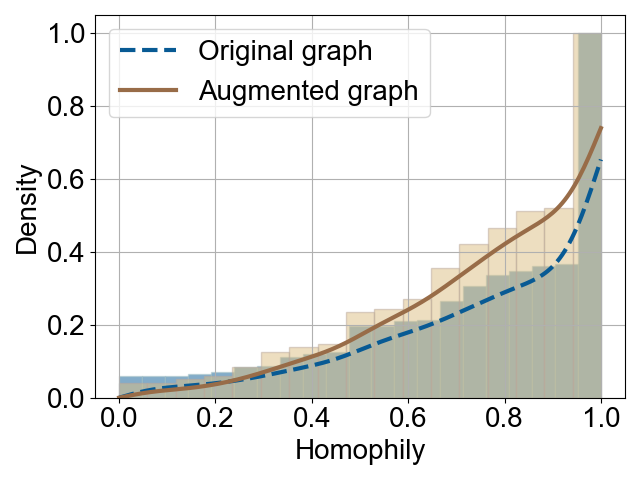}
            }
            \subfigure[Citeseer]{
                \label{subfig:augment_homophily_citeseer}
                \includegraphics[width=0.22\textwidth]{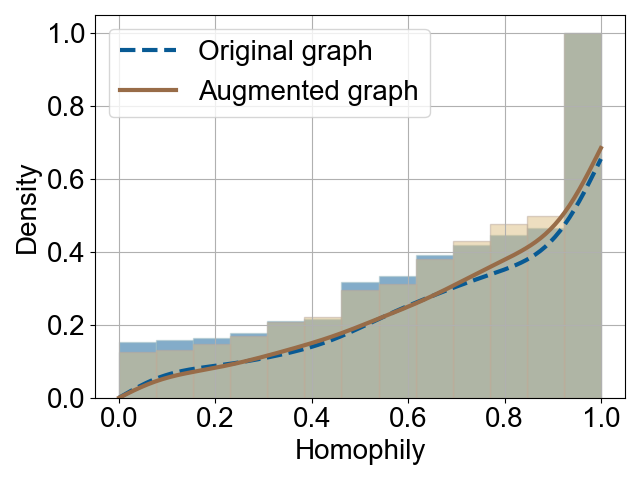}
            }
            \caption{The cumulative distribution of homophily for all nodes in the original and augmented graphs, respectively.}
            \label{fig:augment_homophily}
        \end{figure}

        In this subsection, we aim to understand the augmented graph generated by the tail structure augmentor.
        Specifically, we calculate the homophily of all nodes in both the original and augmented graphs.
        The cumulative distribution of homophily for all nodes in both graphs is presented in \cref{fig:augment_homophily}.
        The horizontal axis represents node homophily, while the vertical axis represents cumulative density, indicating the proportion of nodes with homophily values less than or equal to that value.
        From \cref{fig:augment_homophily}, we observe that there are fewer total-heterophilic nodes in the augmented graph, demonstrating that \ours decreases the proportion of total-heterophilic nodes as expected.
        Additionally, the proportion of nodes with higher homophily values is larger in the augmented graph than in the original graph, confirming that the augmented graph facilitates GNN training by adding pseudo-homophilic edges to tail nodes.

        \subsection{Ablation Study and Parameter Analysis}
        We compare \ours with its four variants to examine the importance of different components.
        Specifically, we implement the following variants:
        (1) \textit{\ours-w/o-Augm.}:
        We employ a vanilla VGAE to conduct neighbor generation and train it with the objective $\mathcal{L} = \gamma \mathcal{L}_{vgae} + \mathcal{L}_{sup}$, where $\mathcal{L}_{vgae}$ is the well-established loss for vanilla VGAE and $\mathcal{L}_{sup}$ is the semi-supervised node classification loss. We fine-tune $\gamma$ for a fair comparison.
        (2) \textit{\ours-w/o-$\mathcal{L}_{aug}$}:
        We remove the augmentation constraint for the tail structure augmentor by setting $\beta$ to zero.
        (3) \textit{\ours-w/o-$\mathcal{L}_{p}$}:
        We remove the propagation constraint for the tail structure augmentor by setting $\eta$ to zero.
        (4) \textit{\ours-w/o-$\mathcal{L}_{ali}$}:
        We remove the alignment constraint for the tail structure augmentor by setting $\delta$ to zero.

        \begin{figure}[t]
            \centering
            \subfigure[Cora]{
                \label{subfig:loss_ablation_cora}
                \includegraphics[width=0.22\textwidth]{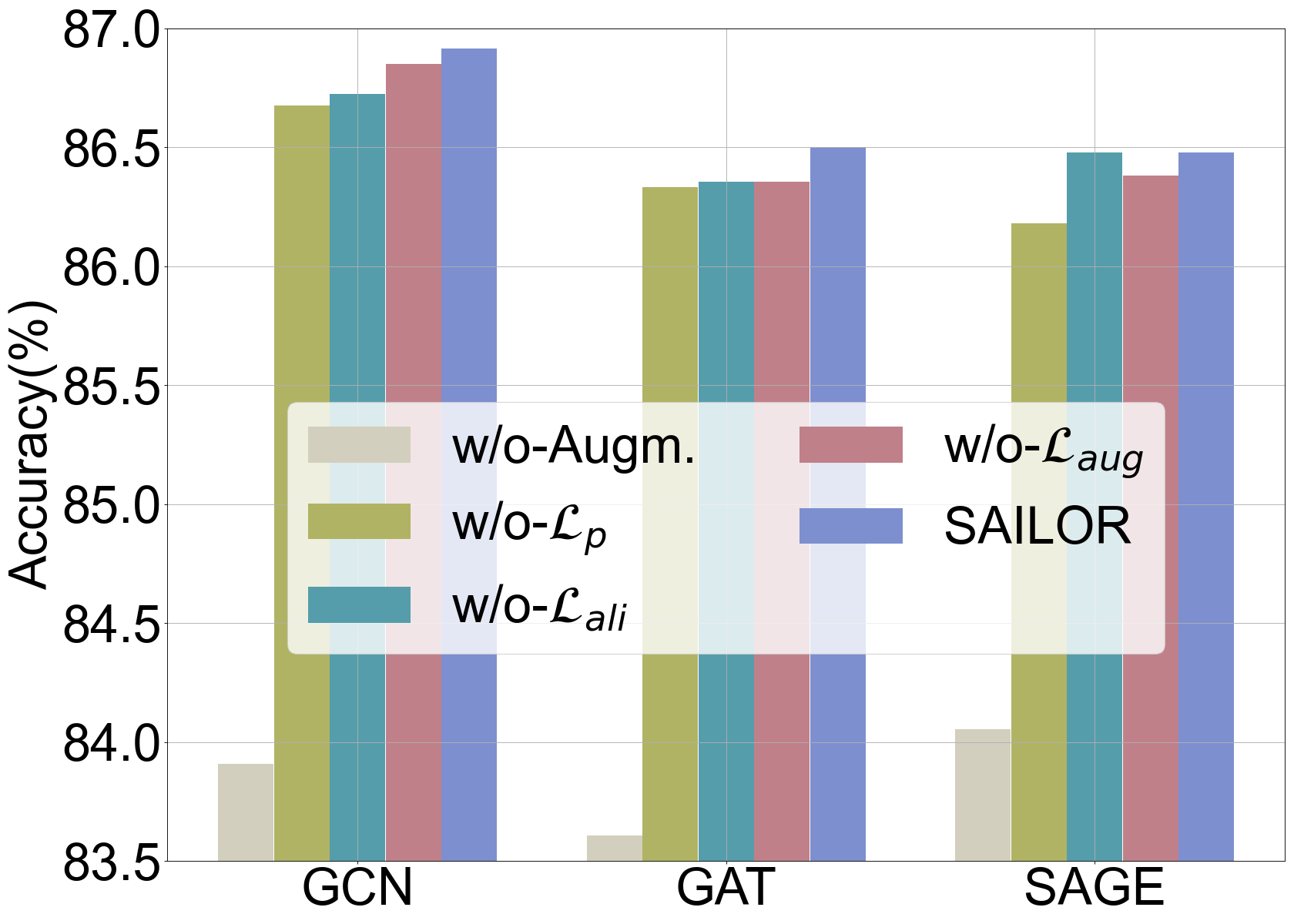}
            }
            \subfigure[Citeseer]{
                \label{subfig:loss_ablation_citeseer}
                \includegraphics[width=0.22\textwidth]{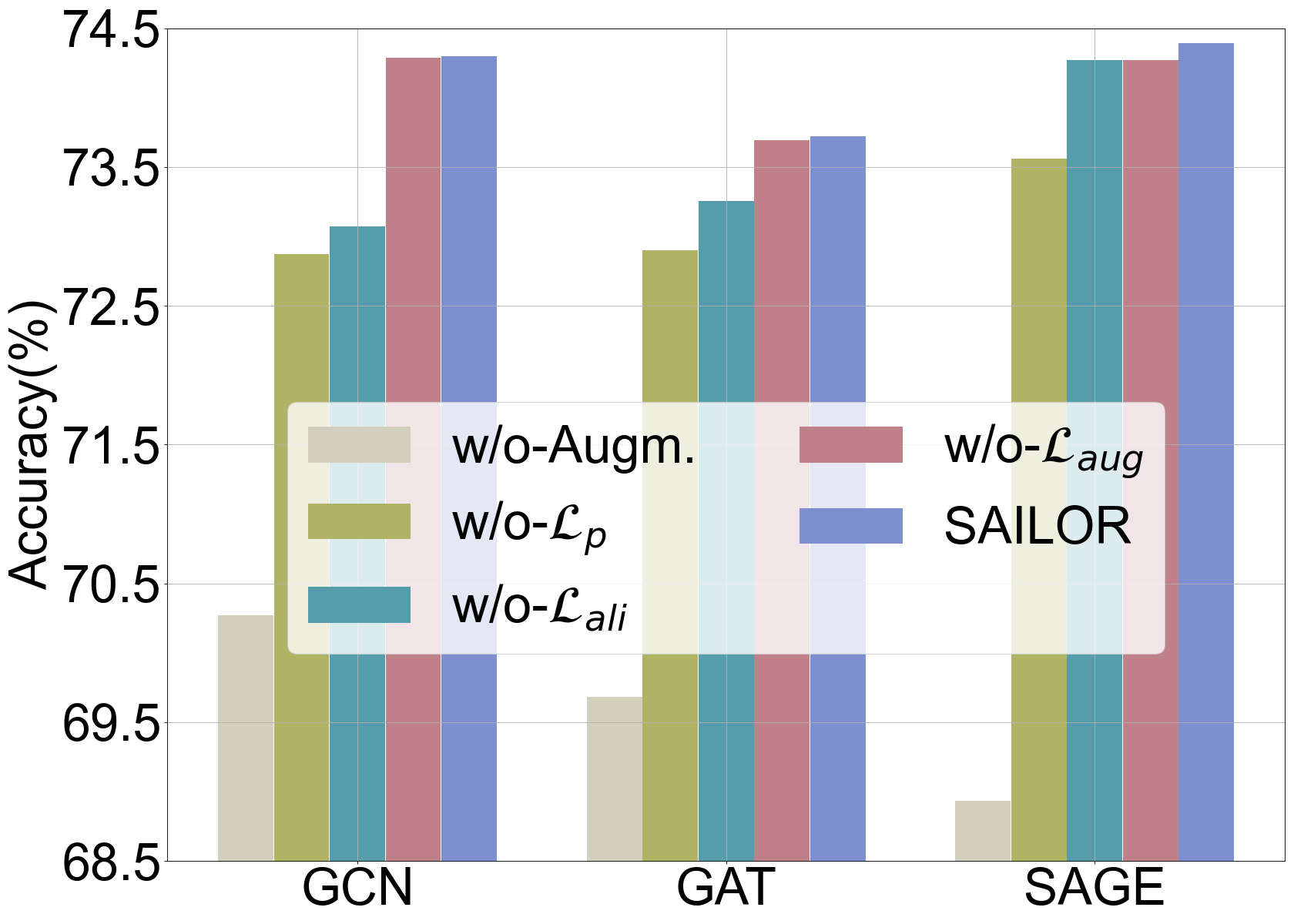}
            }
            \caption{Node classification performance of \ours variants.}
            \label{fig:ablation}
        \end{figure}

        \begin{figure}[t]
            \centering
            \subfigure[Cora]{
                \includegraphics[width=0.22\textwidth]{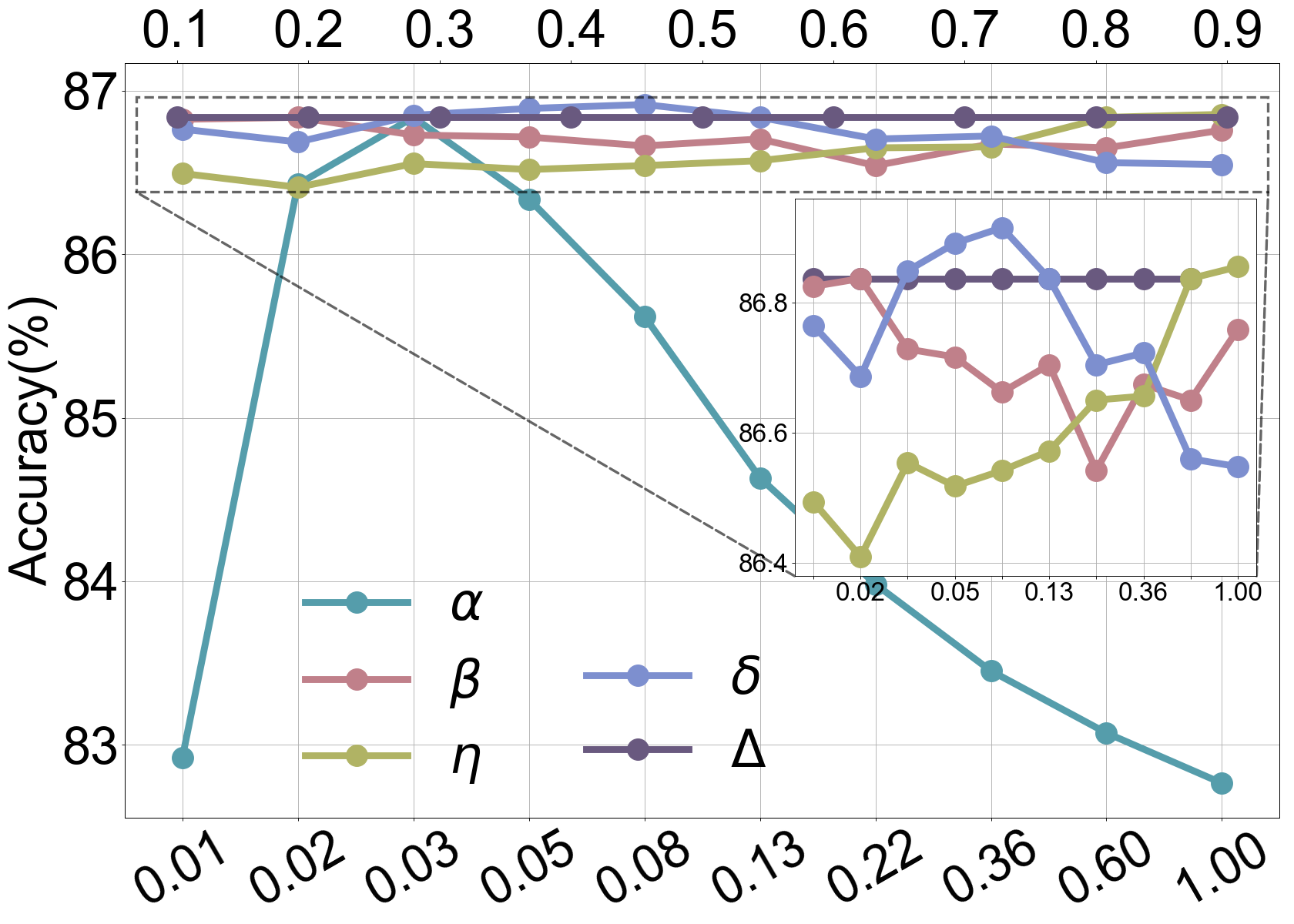}
            }
            \subfigure[Citeseer]{
                \includegraphics[width=0.22\textwidth]{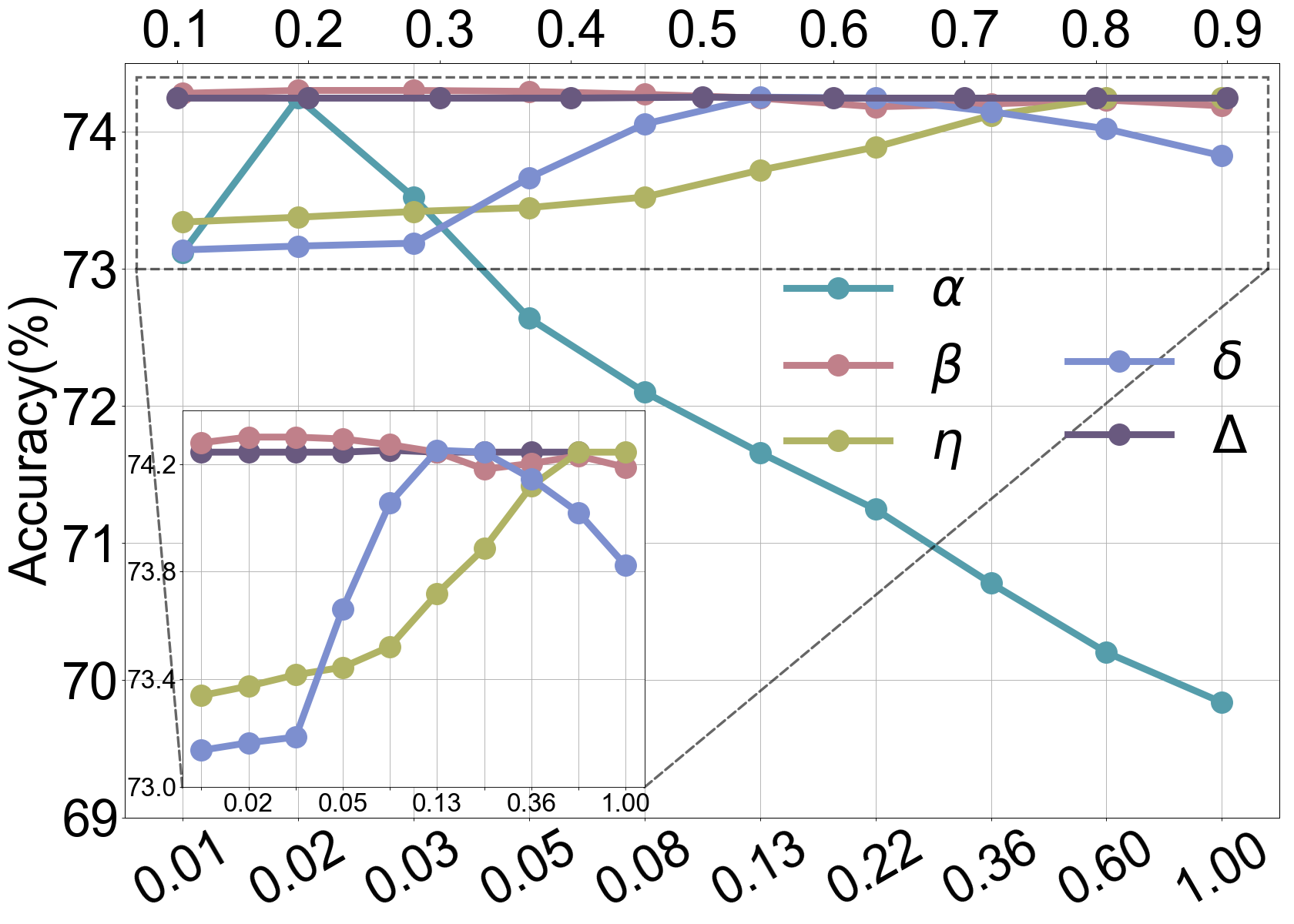}
            }
            \caption{Parameter analysis on Cora and Citeseer.
                The lower horizontal axis represents the values of $\alpha$, $\beta$, $\eta$, and $\theta$, which are evenly spaced on a logarithmic scale from 0.01 to 1.
                The upper horizontal axis represents the values of $\Delta$.}
            \label{fig:params_analysis}
        \end{figure}

        The experiment results are shown in \cref{fig:ablation}, from which we can make several observations.
        Firstly, \textit{\ours-w/o-Augm.} performs significantly worse than all other variants, emphasizing the crucial role of the tail structure augmentor in enhancing the performance of \ours.
        Secondly, all three constraints contribute to the performance of \ours; removing any one of them results in a decrease in performance.
        This indicates that each constraint is necessary for achieving optimal performance.

        We also explored the sensitivity of \ours to various hyper-parameters, including the loss weight of GNN ($\alpha$), the three constraints for the tail structure augmentor ($\beta$, $\eta$, $\delta$), and the edge-dropped proportion for converting head nodes into forged tail nodes ($\Delta$).
        Experiments are performed by altering the value of each hyper-parameter while keeping the others fixed.
        GCN is used as the base architecture for these experiments.
        The results, reported in \cref{fig:params_analysis}, show that changes in the weight of GNN ($\alpha$) can significantly affect the performance of \ours.
        There is an optimal value for $\alpha$, and deviating from this value can impair performance.
        In contrast, the performance of \ours is generally insensitive to changes in $\beta$, $\eta$, $\delta$, and $\Delta$, with performance changes of only about one percentage point at most as their values vary.

\subsection{Tail Node Classification under Public Split}
In this subsection, we align the dataset partition setting with that of GCN \cite{2017ICLR-GCN} to further confirm the effectiveness of \ours.
The experiment results, reported in \cref{tab:public_split}, show that \ours still achieves superior performance in tail node classification under public partitioning setting as in \cite{2017ICLR-GCN}.
Besides, the results indicate that the performance of head nodes has also been significantly improved.
This is because we process the edges in the graph as undirected.
When the tail structure augmentor adds pseudo-homophilic edges to tail nodes, it may also add pseudo-homophilic edges to some head nodes.
This can enhance the structural information of head nodes and promotes their representation learning.

\begin{table}[t]
    \centering
    \caption{Node classification accuracy under public dataset partitioning setting as in \cite{2017ICLR-GCN}.}
    \begin{tabular}{c|ccc}
        \toprule
        \textbf{Citeseer}
         & \textbf{Test}              & \textbf{Head} & \textbf{Tail} \\
        \midrule
        \textbf{GCN}
         & 68.29$\pm$0.65
         & 75.42$\pm$0.52
         & 67.00$\pm$0.78                                             \\
        \textbf{DEMO-Net}
         & 65.48$\pm$1.20
         & 72.35$\pm$1.75
         & 64.24$\pm$1.41                                             \\
        \textbf{Tail-GCN}
         & \underline{69.18}$\pm$1.04
         & \underline{77.78}$\pm$1.65
         & 67.63$\pm$1.01                                             \\
        \textbf{Cold Brew}
         & 66.60$\pm$0.40
         & 74.84$\pm$1.35
         & 65.11$\pm$0.51                                             \\
        \textbf{GRADE}
         & 68.92$\pm$1.37
         & 68.81$\pm$4.61
         & \underline{68.93}$\pm$1.35                                 \\
        \textbf{SAILOR}
         & \textbf{73.91}$\pm$0.74
         & \textbf{80.39}$\pm$2.09
         & \textbf{72.74}$\pm$0.96                                    \\
        \bottomrule
    \end{tabular}
    \label{tab:public_split}
\end{table}

\section{Conclusion}
In this paper, we focus on the problem of tail node representation learning and propose a general structural augmentation based method, \ours.
We find that the lack of structural information in tail nodes is a contributing factor to total-heterophily, which significantly impairs GNNs’ performance.
To tackle this problem, \ours exploits a tail structure augmentor with three well-designed constraint strategies to add pseudo-homophilic edges to tail nodes.
This decreases the ratio of total-heterophilic nodes in the graph, thereby facilitating the training of GNNs.
Extensive experiments on public datasets demonstrate the effectiveness of \ours.

\begin{acks}
    The research is supported by the National Key R\&D Program of China under grant No. 2022YFF0902500, the Guangdong Basic and Applied Basic Research Foundation, China (No. 2023A1515011050), the Tencent AI Lab RBFR2022017.
\end{acks}

\bibliographystyle{ACM-Reference-Format}
\balance
\bibliography{main}










\end{document}